\documentclass[11 pt, onecolumn]{article}
\usepackage[margin=1.1in]{geometry}

\usepackage{amsmath}
\usepackage{bm}	
\usepackage{graphicx}

\usepackage{graphicx}
\usepackage{multirow}
\usepackage{mdwlist}

\usepackage{threeparttable}

\usepackage{amsfonts}
\usepackage{amsmath}
\usepackage{hyperref}
\usepackage{bbm}
\usepackage{float}
\usepackage{caption}
\usepackage{subcaption}
\usepackage{xcolor}
\usepackage{mathtools}
\usepackage{tikz}

\usetikzlibrary{shapes.geometric,arrows}
\usepackage{enumerate}
\usepackage{multicol}
\usepackage{stackrel}

\usepackage{subcaption}
\usepackage{amssymb}
\usepackage[utf8]{inputenc}
\usepackage{booktabs,lipsum}
\usepackage{hyperref}
\usepackage{algpseudocode}
\usepackage{caption}

\usepackage[numbers]{natbib}
\usepackage{longtable}
\usepackage{pdflscape}
\usepackage[ruled,vlined]{algorithm2e}

\DeclareMathAlphabet{\mathcal}{OMS}{cmsy}{m}{n}

\usepackage{graphicx}
\usepackage{epstopdf}
\usepackage{authblk}
\usepackage{float}

\begin{document}
\title{\LARGE \bf Regulator-Manufacturer AI Agents Modeling: Mathematical Feedback-Driven Multi-Agent LLM
Framework}

\author[1]{Yu Han\thanks{These authors contributed equally to this work.}}
\author[2]{Zekun Guo}

\affil[1]{Department of Engineering Science, University of Oxford, United Kingdom \\ \texttt{yu.han@eng.ox.ac.uk}}
\affil[2]{Data Science, Artificial Intelligence and Modelling Center, University of Hull, United Kingdom \\ \texttt{Z.Guo2@hull.ac.uk}}

\newcommand*{\QEDA}{\hfill\ensuremath{\blacksquare}}%

\maketitle

\clearpage
\thispagestyle{empty}
\begin{center}
{\large\bfseries Technical clarification on the implementation and quantitative analyses}
\end{center}

\vspace{0.6em}

\noindent
As part of follow-on work, we re-examined the implementation underlying the
quantitative analyses reported in this paper and revisited the associated
mathematical formulation. This review identified several differences between the
implementation and the methodological description, as well as structural
properties of the model that affect the interpretation of the quantitative
results.

First, the thirteen parameters used in the implementation are fixed values rather
than estimates obtained from data, as described in Section~3.4. The benefit--risk
ratio is also implemented differently from its formal definition in Section~3,
including the placement of the safety score and the use of a fixed threshold. In
addition, the computational configuration used for the analyses in Section~4
differs from that described in the text, and these runs do not involve
language-model calls.

Further examination of the model identified additional properties relevant to
interpretation. In particular, Equation~(2) depends on a positive initial
condition that is not specified in the paper and has a critical guidance
intensity below which compliance effort collapses. The thirteen parameters are
not jointly identifiable from the model outputs, and the parameter settings used
in the implementation fall within a divergent regime. The one-at-a-time
sensitivity analysis also does not account for substantial parameter
interactions.

The mathematical specification in Section~3.2 remains unchanged. However, given
the implementation differences and the additional analysis above, we recommend
that the quantitative results in Section~4, the associated parameter values, and
interpretations based on them not be treated as validated empirical findings or
cited for substantive conclusions.

A complete re-analysis and the corresponding implementation will be made
available with the follow-on work. This note is provided to clarify the status
and limitations of the original quantitative analyses.

\clearpage


\begin{abstract}
The increasing complexity of regulatory updates from global authorities presents significant challenges for medical device manufacturers, necessitating agile strategies to sustain compliance and maintain market access. Concurrently, regulatory bodies must effectively monitor manufacturers' responses and develop strategic surveillance plans. This study employs a multi-agent modeling approach, enhanced with Large Language Models (LLMs), to simulate regulatory dynamics and examine the adaptive behaviors of key actors, including regulatory bodies, manufacturers, and competitors. These agents operate within a simulated environment governed by regulatory flow theory, capturing the impacts of regulatory changes on compliance decisions, market adaptation, and innovation strategies. Our findings illuminate the influence of regulatory shifts on industry behavior and identify strategic opportunities for improving regulatory practices, optimizing compliance, and fostering innovation. By leveraging the integration of multi-agent systems and LLMs, this research provides a novel perspective and offers actionable insights for stakeholders navigating the evolving regulatory landscape of the medical device industry.
\end{abstract}


\section{Introduction}
\subsection{Adaptive Dynamics and Emergent Behaviors in Regulatory Affairs}

The ongoing issuance of guidance documents by global medical products regulatory authorities such as the Food and Drug Administration (FDA) reflects an increasing commitment to public health and safety in response to fast-paced technological advancements. These proactive regulatory actions are aimed at ensuring that new medical devices meet high standards for efficacy and safety. However, as regulatory bodies accelerate the release of new guidelines, medical device manufacturers face mounting pressure to interpret, implement, and comply with updated standards promptly \cite{han2024transforming}. The ability to swiftly adapt to these changes is essential for maintaining market access and ensuring compliance with evolving regulatory expectations \cite{kramer2012does}, \cite{han2024evaluation}.

In regulatory systems, emergent behaviors are evident when new regulations trigger adaptive responses across various stakeholders, including manufacturers, regulatory agencies, and healthcare providers \cite{anderson1999perspective}. These responses can lead to substantial shifts in compliance practices and innovation trajectories. Regulatory interactions are often nonlinear, meaning that small changes in regulations can produce disproportionately large effects on behavior and system-wide outcomes \cite{sterman2000business}. Stakeholders within these systems continually adjust their strategies to align with evolving regulatory standards, highlighting the self-organizing nature of regulatory systems \cite{kauffman1993origins}. Through decentralized decision-making, these systems can lead to the spontaneous emergence of best practices and industry standards \cite{Helbing2013HowTD}.

Despite the importance of understanding these dynamics, a significant gap exists in current research, where most studies focus on regulatory frameworks or compliance outcomes in isolation. This gap leaves essential questions unanswered: How do continuous regulatory updates impact manufacturers' decision-making processes, and what adaptive strategies can support both compliance and innovation in this fast-evolving landscape \cite{lottes2022navigating}?

This study aims to address this research gap by introducing a comprehensive model that uses LLM agents to capture the cyclical interactions between regulatory authorities and medical device manufacturers. By simulating the relationships and feedback loops among regulatory bodies, manufacturers, and the responses triggered when manufacturer agents "review" a set of 10 regulatory guidelines—followed by the regulator agent's decision on whether to approve the submissions—this approach provides an innovative angle within regulatory systems. It also offers practical guidance for both regulatory bodies and manufacturers in developing agile and responsive compliance strategies.

\subsection{Multi-Agent Modeling and Complexity Theory}

Multi-agent modeling offer a powerful approach for simulating complex systems with autonomous, interacting agents \cite{wooldridge2009introduction}. In land-use and land-cover change (LUCC) modeling, MAS combines cellular landscape models with agent-based decision-making representations, effectively capturing spatial interactions and decentralized decision-making \cite{parker2003multi, liu2024multi}. These models are particularly useful for representing heterogeneous conditions and human-environment interactions, ranging from abstract hypotheses to detailed policy analyses. Recent advancements have extended MAS to the multimodal domain, leveraging large language models (LLMs) to create large multimodal agents (LMAs) capable of interpreting and responding to diverse user queries \cite{xie2024large}. LMAs enhance AI agents' ability to handle complex tasks by integrating multiple modalities. However, standardized evaluation methods are needed to facilitate meaningful comparisons among different LMAs and guide future research in this rapidly evolving field \cite{guo2024large}.

Multi-agent modeling (MAM) offers a powerful approach to simulate complex systems comprising multiple autonomous agents with interactive behaviors. This strength is particularly suited for examining dynamic and emergent properties in systems where agents—such as manufacturers and regulatory bodies, continuously adjust their behaviors based on interactions with one another and with their environment \cite{turner2019complexity}. This bottom-up approach enables researchers to study how localized actions and interactions scale into broader system-wide outcomes, making MAM an effective tool for understanding regulatory environments characterized by unpredictability and non-linearity \cite{macal2016everything}.

At the foundation of multi-agent modeling lies Complexity Theory, which provides a framework for analyzing systems where interconnected components exhibit emergent behaviors. Complexity Theory suggests that even small regulatory adjustments can lead to disproportionately large shifts in stakeholder behavior, highlighting the importance of adaptability within regulatory frameworks \cite{axelrod2008harnessing}. In regulatory systems, such adaptability materializes through a continuous cycle of response, where stakeholders dynamically adjust to regulatory changes, and these adjustments subsequently influence the development of future standards through feedback loops.

Regulatory Flow Theory, which contextualizes MAM in this study, visualizes regulatory interactions as currents within a river, guiding stakeholders' actions. This theory conceptualizes the regulatory landscape as a continuous flow, where each regulatory update acts as a current shaping compliance behavior, strategic planning, and industry innovation. Regulatory Flow Theory captures the cyclical nature of these systems: an initial regulatory document prompts adaptive actions by manufacturers, who then provide feedback that informs further updates. This feedback mechanism creates an evolving regulatory environment, enabling both regulators and manufacturers to respond flexibly to emerging challenges \cite{hedstrom2010causal, wang2024reinforcement}. For regulatory bodies, recognizing the emergent properties of these systems can inform the development of more adaptive and responsive frameworks. For manufacturers, appreciating the complex interactions within the regulatory landscape can enhance strategic planning, enabling them to remain agile and compliant amidst regulatory shifts. This research models the regulatory affairs process, including the issuance of regulations, manufacturers' adaptive compliance efforts, and the influence of stakeholder feedback on regulatory authorities. Ultimately, it aims to explore the balance of delivering safe medical devices to patients without stifling innovation—a nuanced challenge underexplored within Regulatory Sciences. 
\section{Literature Review}

Regulatory authorities are responsible for establishing regulations and guidelines that define new methods, standards, and models to ensure that medical products meet stringent criteria for safety, efficacy, quality, and performance. The field of regulatory affairs is pivotal in overseeing these medical products, working to protect public health by verifying that all products brought to market are both safe and effective \cite{turner2011editor}. Situated at the intersection of engineering and medicine, regulatory science addresses the increasingly complex challenges associated with medical product innovation and oversight \cite{han2024more}. Over the past few decades, the regulatory landscape for medical products has evolved significantly \cite{han2024evaluation}, \cite{han2024revolutionizing}. This transformation has been driven by the need to strike a delicate balance between maintaining high safety standards and fostering innovation to accommodate rapid technological advancements. To advance regulatory practices, researchers are exploring innovative approaches that span from the optimization of language and communication strategies \cite{han2024use} to the integration of artificial intelligence and advanced modeling techniques \cite{han2024transforming}. These advancements aim to enhance regulatory efficiency, improve the interpretation of complex guidelines, and facilitate adaptive regulatory frameworks that keep pace with emerging technologies.


In the field of Regulatory Science, previous modeling efforts have primarily focused on evaluating the safety and effectiveness of individual categories of medical products. Typically, these models are developed for a specific type of pharmaceutical product that shares a common target or mechanism of action, requiring a high degree of precision tailored to the unique characteristics of that product category. Examples include models like the Physiologically Based Pharmacokinetic (PBPK) Absorption Model for oral capsules \cite{stillhart2019pbpk}, clinical decision support systems integrated into medical devices \cite{morrison2018advancing}, and models that leverage real-world evidence throughout the entire product lifecycle \cite{o2019ispor}. Additionally, novel analytical technologies have been developed to improve the detection and assessment of medical products \cite{wang2018analytical}.

From a broader process perspective, considering the development of medical products and their impact on global public health, various other types of models have been designed to streamline regulatory processes and improve compliance. A foundational model in this domain is the risk management Model, which gained prominence following the introduction of the ICH Q9 guidelines. This model emphasizes the identification, assessment, and mitigation of risks throughout both the drug development and post-market phases, offering a framework that aligns regulatory oversight with public safety objectives \cite{ICH2005}, \cite{fda2006guidance}. Research has shown that risk-based regulation enhances regulatory outcomes by prioritizing resources and focusing on the most critical risk factors \cite{gonccalves2020risk}.

Another pillar of modeling in Regulatory Science is the Lifecycle Management Model, which conceptualizes drug regulation as a continuous process encompassing development, market approval, and post-market surveillance. This model's strength lies in its ability to uphold drug quality and safety across the entire product lifecycle, adapting to emerging concerns through responsive and flexible regulatory mechanisms \cite{smith2017structured}. Additionally, economic analysis serves as another pillar, with the Cost-Benefit Analysis Model providing a quantitative method to evaluate the economic impacts of new regulations. This approach allows regulators to balance public health benefits against the financial costs to the pharmaceutical industry, fostering rational and balanced decision-making \cite{weinstein1977foundations}, \cite{drummond2015methods}.


However, in the field of regulatory science, dynamic modeling that captures the intricate relationships among regulators, manufacturers, and the regulatory content they release remains largely underexplored. Understanding how manufacturers respond to regulatory guidelines and adapt their practices is crucial for optimizing compliance and fostering innovation. To address this gap, Regulatory Flow Theory, a novel model supported by the Transcendence Equation, has been introduced. This model mathematically represents the feedback and adaptation mechanisms among key stakeholders in the regulatory landscape. Specifically, the equations describe how key variables such as regulatory influence, compliance effort, market adaptation, and feedback interact over time. By formalizing these relationships, the model provides a nuanced understanding of how regulatory decisions impact industry behavior and compliance strategies.


The development for such dynamic equation was historically hindered in the past, primarily due to the difficulty in capturing and experimentally validating the intricate relationships. However, with the advent of Large Language Models (LLMs), which significantly expanded the potential for dynamic modeling. Integrating LLMs within multi-agent frameworks \cite{wooldridge2009introduction} offers a promising approach for optimizing complex engineering systems. In these frameworks, agents are assigned distinct roles, and through role-playing interactions, they simulate the behaviors of various entities. The multi-agent system consists of numerous interacting agents that collaborate or compete to achieve specific objectives \cite{xiong2019analysis}.

Multi-Agent Models have the potential to revolutionize regulatory processes by enabling autonomous agents, each representing a different stakeholder or regulatory entity, to interact and make decisions collaboratively. These agents, equipped with LLM capabilities, can interpret regulatory texts, process real-time data, and respond dynamically to changes in the regulatory environment. Moreover, the use of mathematical feedback mechanisms within these models allows for more adaptive and responsive decision-making, a critical advancement in managing complex regulatory ecosystems. The promise of multi-agent systems in regulatory contexts lies in their ability to facilitate collaboration, enhance real-time decision-making, and scale to accommodate diverse regulatory requirements. By leveraging LLMs for feedback-driven optimization, these models can deliver predictive insights previously unattainable, effectively addressing emerging regulatory challenges in ways that traditional approaches cannot \cite{bao2024dynamic}. The potential for such models to transform regulatory practices is substantial, offering a more proactive, data-driven, and adaptable framework for oversight, particularly in the medical industry.

\section{Methodology} \label{Section: model}

Our study employed Large Language Models (LLMs) to simulate the behavior of manufacturer agents in response to regulatory prompts, systematically documenting each agent’s strategic decisions, adaptations, and overall performance \cite{gao2024empowering}. We examined how agents navigated regulatory challenges, interpreted new guidelines, and adjusted their product designs to meet evolving compliance requirements. The evaluation framework was modeled using a Benefit-Risk Ratio (BRR) assessment \cite{kurzinger2020structured}, inspired by FDA methodologies for evaluating product safety and effectiveness. At each time step, we calculated BRR values to assess whether agents met regulatory standards, analyzing their reasoning and adaptive strategies in response to regulatory pressure.

The BRR calculation begins by extracting essential scores from each manufacturer agent’s submission. These include a safety score, an effectiveness score, and a compliance score, each rated on a scale from 1 to 10. The compliance score is derived directly from the agent’s decision rating, which reflects the extent to which the product aligns with regulatory guidelines. To compute the overall benefit, the model sums the safety, effectiveness, and compliance scores, representing a holistic measure of the product’s positive impact. The risk component, represented by the adverse event score, quantifies potential negative outcomes associated with the product. The BRR is then calculated as the ratio of the aggregated benefit to the adverse event score, providing a comprehensive metric for regulatory evaluation. The final decision on whether to approve or reject a submission is determined by comparing the calculated BRR with a predefined approval threshold. If the BRR meets or exceeds this threshold, the product is deemed acceptable and receives approval. Conversely, if the BRR falls below the threshold, the submission is rejected. This structured approach allows the model to simulate real-world regulatory processes, similar to those employed by agencies such as the FDA, and ensures that the assessment balances safety, efficacy, and regulatory compliance in a robust and replicable manner\cite{kurzinger2020structured}.

\begin{figure}[h]
    \centering
\includegraphics[width=1\textwidth]{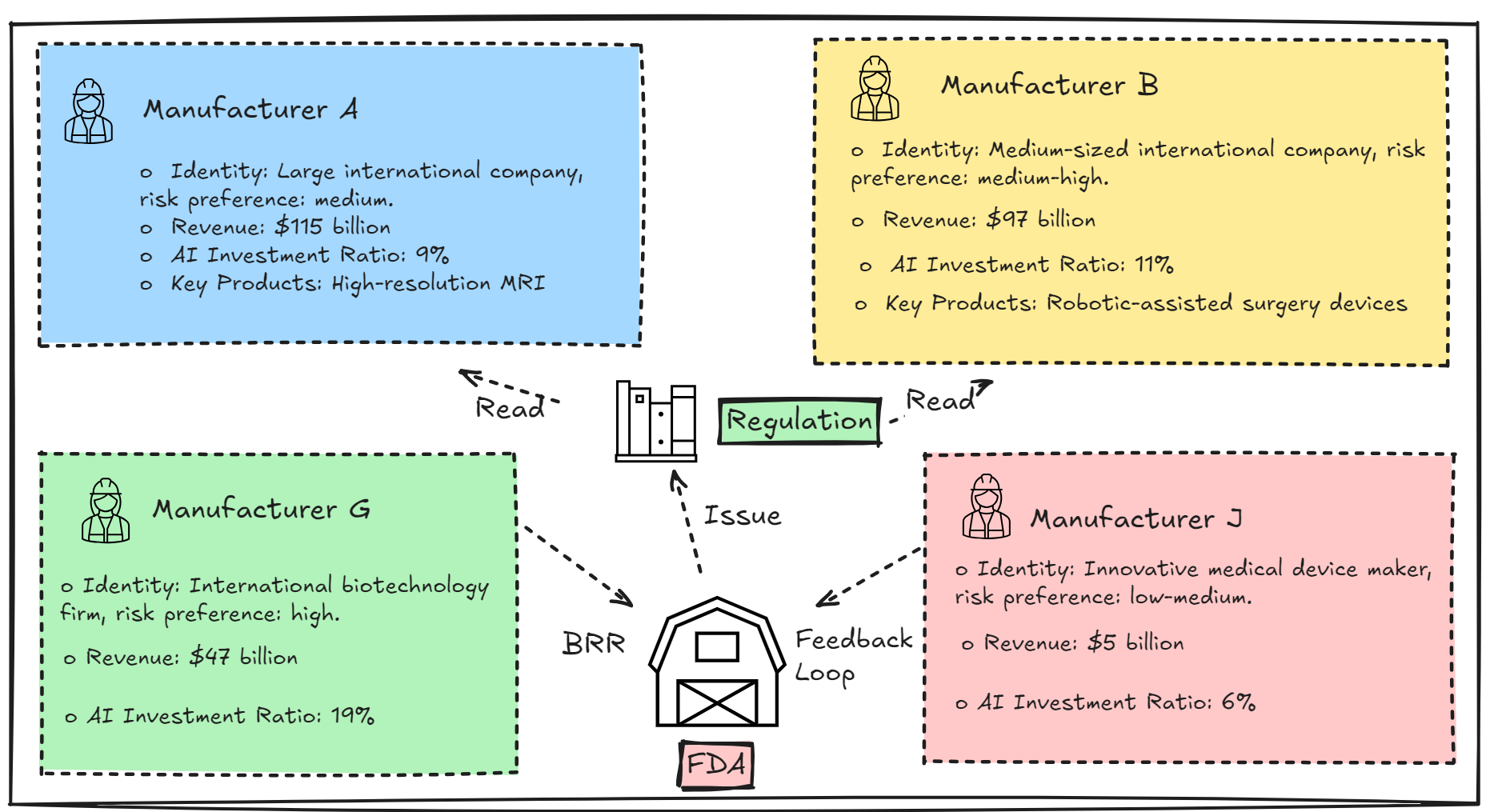}
    \caption{Interaction Between Regulatory Authority and Manufacturers Agents in Approval Processes}
    \label{fig:1}
\end{figure}

We developed a comprehensive simulation environment comprising 10 distinct manufacturer agents and one regulatory agent, designed to emulate a dynamic regulatory ecosystem. The regulatory agent issued new guidelines, prompting the manufacturer agents to adapt and respond strategically. We created a set of 10 regulations—5 strict and 5 lenient—and exposed agents to strict regulations for 10 time steps, followed by lenient regulations for 5 time steps, recording their responses throughout, as shown in Figure \ref{fig:1}.

\begin{figure}[h]
    \centering
\includegraphics[width=1\textwidth]{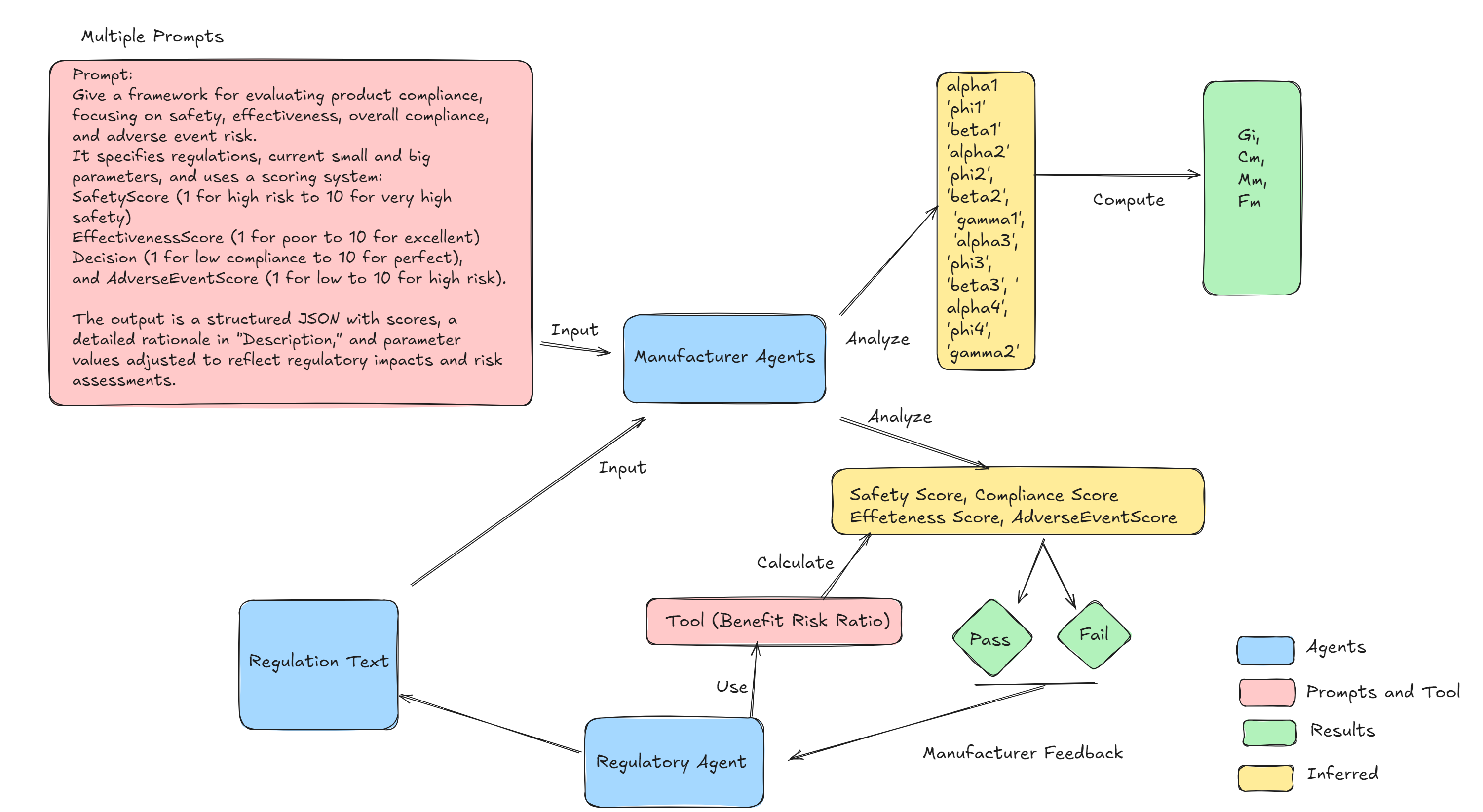}
    \caption{Agent-Based Simulation of Pharmaceutical Regulatory Review Process}
    \label{fig:2}
\end{figure}

Agents were required to make decisions based on several factors, including the regulatory environment, financial resources, AI strategies, and plans for market adaptation. They also had to determine which products they could feasibly develop under these conditions. These decisions were then evaluated by the regulatory agent, using a risk-benefit ratio calculator to determine compliance. The calculator integrated metrics such as safety, effectiveness, and compliance, providing a quantitative framework for decision-making. Each metric, scored from 1 to 10, was aggregated to calculate the benefit, which was then divided by the adverse event score to yield the BRR. A submission was approved if the BRR met or exceeded a predefined threshold; otherwise, it was rejected. This approach ensured a systematic, balanced assessment of safety, efficacy, and regulatory adherence, reflecting real-world practices as shown in Figure \ref{fig:2}.

The simulation embedded regulatory knowledge and mathematical frameworks within the agents' decision-making processes, guiding even those without a comprehensive understanding of device regulation to make informed choices. By modeling these interactions through nonlinear differential equations, we gained insights into how regulatory pressures influence compliance and innovation in the AI medical device sector. This approach allowed us to capture the complex dynamics between regulatory guidance and market responses, providing a detailed analysis of adaptive behaviors over time.

\subsection{Model Framework and Initialization}

Our experiment is built upon a two-layer design framework, integrating both mathematical modeling and multi-agent LLM simulations to rigorously investigate the interplay between regulatory guidance and manufacturer behavior. The first layer of this design involves constructing a set of transcendental equations that form the core of a realistic physical world model. We developed four key equations to describe the process by which a regulatory authority, such as a pharmaceutical or medical device regulatory body, issues guidance and how manufacturer agents adapt to this evolving regulatory environment. These equations capture the intricate feedback loops and adaptive behaviors that characterize real-world regulatory dynamics.

The four main variables in this model are $G_i$, $C_m$, $M_m$, and $F_m$. Specifically, $G_i$ represents the guidance issuance rate, reflecting how quickly new regulatory requirements are introduced. $C_m$ denotes the compliance effort, a measure of the resources and actions manufacturers allocate to meet regulatory standards. $M_m$ stands for market adaptation, illustrating how well manufacturers adjust their strategies and operations in response to regulatory changes. Lastly, $F_m$ captures the feedback from manufacturers, representing the influence of collective industry responses on future regulatory decisions. Together, these equations simulate the complex interactions between regulatory authorities and manufacturers, offering a mathematically grounded view of compliance dynamics.

To ensure the equations accurately represent real-world scenarios, we apply a nonlinear least squares optimization method to determine initial values for the parameters, such as $alpha_1$, $phi_1$, $beta_1$, and others. This optimization aligns the model's baseline state with either observed or theoretical expectations, providing a high-fidelity starting point for our simulations. Once we have these optimized parameters, we calculate the initial values for $G_i$, $C_m$, $M_m$, and $F_m$. These initial conditions set the stage for a realistic simulation, representing each agent's starting point in terms of regulatory compliance and market positioning.

The second layer of our framework introduces multiple agents using large language models (LLMs) to simulate how manufacturer agents process regulatory guidance with Camel AI platform \cite{li2023camel}. Each agent reads and interprets regulatory documents through carefully crafted prompts, which simulate the decision-making process. Based on their understanding, the agents strategically adjust their internal parameters. For instance, an agent might increase $alpha_1$ to accelerate guidance issuance if the regulations demand stricter compliance, or they might adjust $phi_2$ to fine-tune their research and development investments in line with regulatory requirements. The reasoning behind each adjustment is recorded, providing a rich narrative that reveals the strategic thought processes of the agents.

A highlight of our approach is the dynamic interaction between the mathematical model and the LLM-based agents. After each agent makes its decisions and adjusts its parameters, these new values are fed back into the transcendental equations. The updated equations are then solved to calculate new values for $G_i$, $C_m$, $M_m$, and $F_m$. This iterative process creates a feedback loop where the mathematical model informs the LLM agents, and the agents’ decisions refine the mathematical outcomes. By grounding LLM decisions in a mathematical framework, we enhance the precision and realism of the simulations.

This integration ensures that the LLM agents' strategic decisions are not only guided by linguistic understanding but are also validated and refined through rigorous mathematical calculations. As a result, the model becomes more robust, capturing the nuances of regulatory compliance and adaptation in a way that is both theoretically sound and practically relevant. This innovative combination of mathematical modeling and LLM simulations provides a comprehensive platform for exploring how manufacturers might respond to changing regulatory landscapes, offering insights into the effectiveness of different regulatory strategies.

The experiment is grounded in a set of nonlinear differential equations that model key variables: Guidance Issuance Rate ($G_i$), Compliance Effort ($C_m$), Market Adaptation ($M_m$), and Manufacturer Feedback ($F_m$). The experiment proceeds as follows:

\begin{algorithm}[H]
  \SetAlgoNlRelativeSize{-1} 
  \SetAlgoLined
  \SetKwInput{KwIn}{Input} 
  \SetKwInput{KwOut}{Output}
  
  \KwIn{Initial state variables for each manufacturer agent; Total time steps $T$; System parameters}
  \KwOut{Final state variables after simulation}
  
  \BlankLine
  \textit{Module 1: Initialization}
  
  Initialize manufacturer agents with different profiles, represented as: 
  
  \textit{ManufacturerAgents} = \{10 instinct agents with each identity card\}
  
  \BlankLine
  \ForEach{agent in \textit{ManufacturerAgents}}{
      Initialize agent's state variables $G_i$, $C_m$, $M_m$ \\
      Assign agent.level based on corresponding profile characteristics
  }
  
  \BlankLine
  \textit{Module 2: Main Simulation Loop}
  
  \For{$t \gets 1$ \KwTo $T$}{
      \BlankLine
      \textit{Step 1: Regulatory agent issues new regulations} \\
      regulations $\gets$ GenerateRegulations$(t)$
      
      \BlankLine
      \textit{Step 2: Manufacturers read regulations and prepare for compliance}
      
      \ForEach{agent in \textit{ManufacturerAgents}}{
          \textit{Read and interpret regulations}
      }
      
      \BlankLine
      \textit{Step 3: Manufacturers decide on compliance and update states}
      
      \ForEach{agent in \textit{ManufacturerAgents}}{
          \BlankLine
          \textit{3.1 Generate prompt reflecting agent level} \\
          prompt $\gets$ GeneratePrompt(agent.level, regulations)
          
          \BlankLine
          \textit{3.2 Agent uses decision model to decide on compliance} \\
          decision $\gets$ Decision(agent, prompt, model)
          
          \BlankLine
          \textit{3.3 Update state variables based on decision} \\
          $G_i$, $C_m$, $M_m$, $F_m$ $\gets$ UpdatePhysicalState(agent, decision, params)
          
          \BlankLine
          \textit{3.4 Calculate compliance cost and market adaptation} \\
          cost, market\_adaptation $\gets$ CalculateCostAndAdaptation(agent, params)
          
          \BlankLine
          \textit{3.5 Agent submits compliance documents} \\
          \uIf{decision == "Comply"}{
              submission $\gets$ PrepareSubmission(agent, regulations) \\
              BRR $\gets$ CalculateBRR(submission) \\
              outcome $\gets$ EvaluateSubmission(BRR, approval\_threshold)
          }
          
          \BlankLine
          \textit{3.6 Update agent's state variables} \\
          agent.state $\gets$ \{ $G_i$, $C_m$, $M_m$, $F_m$ \} \\
          UpdateAgentState(agent, cost, market\_adaptation)
      }
      
      \BlankLine
      \textit{Step 4: Regulatory agent reviews submissions and updates environment} \\
      feedback $\gets$ CollectFeedback(submissions, outcomes) \\
      UpdateEnvironment(feedback)
      
      \BlankLine
      \textit{Step 5: Record agents’ reactions and analyze results} \\
      RecordAndAnalyzeReactions(agent.states)
  }
  
  \BlankLine
  \textit{Module 3: Output Results} \\
  Output all state variables and analysis results
\end{algorithm}

Our approach begins by optimizing the initial parameters ($\alpha_1, \phi_1, \beta_1$, etc.) through nonlinear least squares, ensuring a strong foundation for simulations. We then compute initial values for $G_i, C_m, M_m, F_m$, establishing baseline states for each manufacturer agent.  Agents interpret regulatory prompts, updating parameters based on strategic reasoning, such as adjusting compliance efforts or market adaptation. These updates are informed by a structured decision-making process and recalculated into new variable states. The simulation includes scenario-based role-playing, where agents adapt further as the narrative evolves, making strategic adjustments documented with rationale. We collect and visualize data at every step, mapping parameter changes and variable trends. This analysis highlights the impact of regulatory guidance and agent interactions, revealing patterns and strategic differences in compliance and market adaptation.

\subsection{1st Mathematical Layer}

Our model conceptualizes the regulatory environment as an adaptive flow, similar to a river shaping the landscape it traverses. Regulatory releases act as “currents” that influence the behavior of manufacturers. Using principles of Multi-Agent Modeling (MAM), we simulate a continuous regulatory-reaction cycle where feedback and regulatory updates interact dynamically. The system of equations presented below captures how regulatory guidance and feedback shape compliance efforts and market adaptation over time.

\begin{align}
\frac{dG_i(t)}{dt} &= \alpha_1 \cdot (1 - e^{-\phi_1 \cdot t}) - \beta_1 \cdot F_m(t) \label{eq:1} \\
\frac{dC_m(t)}{dt} &= \alpha_2 \cdot G_i(t) \cdot \left(1 - e^{-\phi_2 \cdot C_m(t)}\right) - \beta_2 \cdot \left( \frac{C_m(t)}{1 + \gamma_1 \cdot M_m(t)} \right) \label{eq:2} \\
\frac{dM_m(t)}{dt} &= \alpha_3 \cdot C_m(t) \cdot \left(1 - e^{-\phi_3 \cdot G_i(t)}\right) - \beta_3 \cdot M_m(t) \label{eq:3} \\
F_m(t) &= \alpha_4 \cdot \left( \frac{M_m(t) \cdot \left(1 - e^{-\phi_4 \cdot C_m(t)}\right)}{1 + \gamma_2 \cdot C_m(t)} \right) \label{eq:4}
\end{align}

The equation for the rate of guidance issuance, $\frac{dG_i(t)}{dt}$, reflects a balance between natural growth and inhibitory feedback. The term $\alpha_1 \cdot (1 - e^{-\phi_1 \cdot t})$ models an initial rapid increase in guidance issuance that slows over time, akin to the Logistic Growth Model \cite{verhulst1838law}. The inhibitory term $- \beta_1 \cdot F_m(t)$ aggregates feedback from manufacturers, aligning with systems dynamics principles where growth is moderated by systemic constraints \cite{strogatz2018nonlinear}.

Compliance efforts, $\frac{dC_m(t)}{dt}$, are driven by the influence of guidance issuance and market adaptation. The term $\alpha_2 \cdot G_i(t) \cdot \left(1 - e^{-\phi_2 \cdot C_m(t)}\right)$ quantifies how compliance scales with guidance, akin to innovation diffusion models \cite{rogers2003diffusion}. Constraints are modeled by $- \beta_2 \cdot \left( \frac{C_m(t)}{1 + \gamma_1 \cdot M_m(t)} \right)$, capturing the diminishing returns on compliance as market adaptation grows \cite{forrester1968principles}.

Market adaptation, $\frac{dM_m(t)}{dt}$, depends on compliance efforts and guidance issuance. The compounding effect is captured by $\alpha_3 \cdot C_m(t) \cdot \left(1 - e^{-\phi_3 \cdot G_i(t)}\right)$, which slows as saturation occurs, modeled by $- \beta_3 \cdot M_m(t)$ \cite{strogatz2018nonlinear}.

The feedback factor $F_m(t)$ is modeled as $\alpha_4 \cdot \left( \frac{M_m(t) \cdot \left(1 - e^{-\phi_4 \cdot C_m(t)}\right)}{1 + \gamma_2 \cdot C_m(t)} \right)$, representing how market adaptation and compliance efforts influence feedback, incorporating diminishing returns \cite{murray2002mathematical}.

This mathematical framework captures the complex interplay between regulatory guidance, compliance dynamics, and feedback effects, providing a structured approach to modeling regulatory compliance and market adaptation.

\subsection*{Interdependencies and System Dynamics}

The interdependencies in our model illustrate the regulatory system's complexity, emphasizing the interconnected nature of feedback, compliance, and market adaptation.

\begin{itemize}
  \item \textbf{Feedback ($F_m(t)$)} directly influences the \textbf{Guidance Issuance Rate ($G_i(t)$)} through the inhibitory term $- \beta_1 \cdot F_m(t)$, modulating the speed at which new regulatory guidance is issued.
  
  \item \textbf{Guidance Issuance Rate ($G_i(t)$)} impacts \textbf{Compliance Efforts ($C_m(t)$)} via the term $\alpha_2 \cdot G_i(t) \cdot \left(1 - e^{-\phi_2 \cdot C_m(t)}\right)$, driving manufacturers to allocate resources toward meeting regulatory standards.
  
  \item \textbf{Compliance Efforts ($C_m(t)$)} shape \textbf{Market Adaptation ($M_m(t)$)} through the term $\alpha_3 \cdot C_m(t) \cdot \left(1 - e^{-\phi_3 \cdot G_i(t)}\right)$, facilitating strategic adjustments to maintain competitive positioning.
  
  \item \textbf{Market Adaptation ($M_m(t)$)} and \textbf{Compliance Efforts ($C_m(t)$)} collectively influence \textbf{Feedback ($F_m(t)$)} through $\alpha_4 \cdot \left( \frac{M_m(t) \cdot \left(1 - e^{-\phi_4 \cdot C_m(t)}\right)}{1 + \gamma_2 \cdot C_m(t)} \right)$, which reflects how market and compliance dynamics inform the regulatory environment.
\end{itemize}

  \begin{figure}[h]
    \centering
\includegraphics[width=0.9\textwidth]{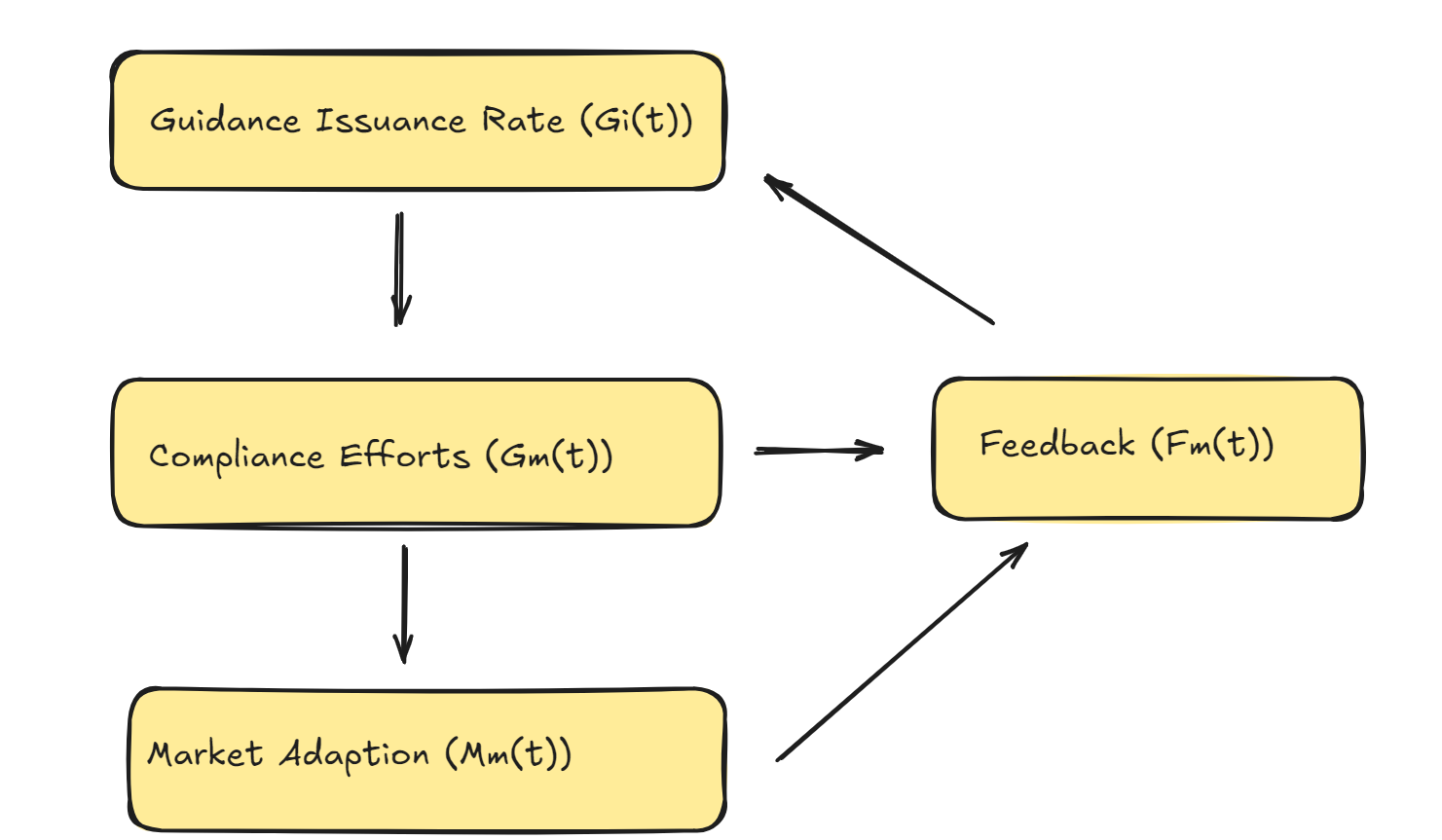}
    \caption{Diagram of interdependencies among variables, illustrating the regulatory feedback loop.}
    \label{fig:loop}
\end{figure}

The model depicts a cohesive regulatory flow: Feedback from manufacturers influences the Guidance Issuance Rate, which determines how rapidly new regulations are introduced. This rate subsequently drives Compliance Efforts, compelling manufacturers to align with regulatory expectations. Compliance Efforts, in turn, catalyze Market Adaptation, enabling strategic adjustments to maintain relevance. Finally, Market Adaptation and Compliance Efforts shape the Feedback mechanism, which informs future regulatory strategies, closing the adaptive loop.

\subsection{2nd Agents Role Play}

\textbf{Regulatory Bodies}
These agents represent regulatory authorities responsible for issuing guidance documents, updating standards, and overseeing compliance. Operating on a cyclical basis, regulatory bodies periodically release new guidelines to reflect the changing landscape of medical device standards, influenced by emerging technologies, public health data, and industry feedback. These updates act as "currents" within the regulatory flow, driving adaptive responses from other agents.

\textbf{Manufacturers}
These agents simulate medical device manufacturers, tasked with interpreting new guidelines, assessing compliance costs, and adjusting operational strategies to maintain market access. Each manufacturer agent is designed with adaptive algorithms that allow them to allocate resources efficiently, balance innovation with regulatory demands, and anticipate future changes. Their strategic adjustments in response to new regulations highlight the self-organizing nature of regulatory systems, where decentralized decisions result in industry-wide compliance patterns.

\textbf{Competitors (Other Manufacturers)}
Competitors within the industry engage in adaptive strategies to maintain competitive positions while responding to regulatory changes. By monitoring regulatory developments and adapting their compliance approaches, they create an environment of continuous innovation and improvement. This inter-competitive dynamic reinforces the necessity for manufacturers to stay agile, balancing regulatory adherence with the need to maintain market share.


\subsection{Environmental Setup}
The simulation environment models the regulatory landscape as a dynamic flow, where guidance updates act like currents, continuously influencing and reshaping agent behaviors. Regulatory bodies release guidance documents at regular intervals, informed by compliance data and feedback from manufacturers and healthcare providers. This steady stream of regulatory information drives manufacturers to adapt their compliance strategies, ensuring they remain aligned with evolving standards. Feedback mechanisms play a crucial role, allowing manufacturers and healthcare providers to share insights and responses with regulatory authorities, which, in turn, shape future updates. This creates a continuous cycle where real-world experiences and regulatory guidance influence each other. Manufacturers must adapt by balancing their resource allocation and compliance efforts, mirroring the real-world challenges and ongoing evolution of the regulatory environment.

\begin{figure}[h]
    \centering
    \includegraphics[width=1\textwidth]{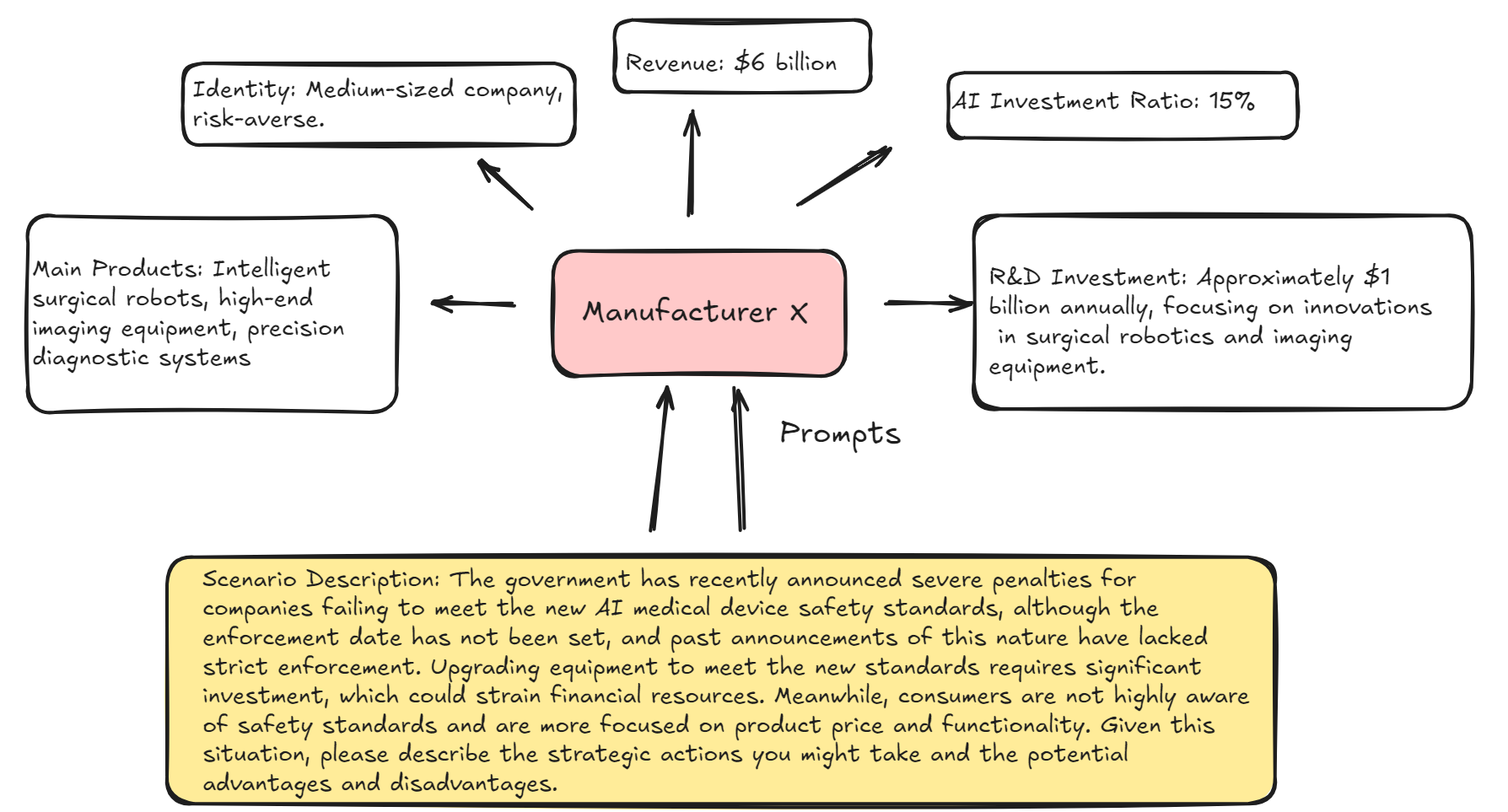}
    \caption{Prompts Example}
    \label{fig:manu}
\end{figure}

We developed comprehensive profiles for our manufacturer agents, each with unique attributes to reflect real-world variability. As illustrated in Figure \ref{fig:manu}, Company B, for instance, has a specific set of characteristics that influence its strategic decision-making. Across our simulation, all 10 agents are designed with distinct profiles, enabling nuanced interpretations and informed responses to regulatory requirements. To simulate regulatory challenges, we employed natural language processing (NLP) techniques to extract five critical topics from FDA regulatory texts pertaining to AI medical devices. We compiled these into a streamlined regulatory guide, abstracting complex guidelines into clear, actionable directives for agents to process efficiently. These scenarios were then deployed on the Camel AI platform, facilitating dynamic interactions among agents as they adapt to evolving regulatory landscapes.

The Camel AI environment supports parallel execution of agents, adaptive feedback mechanisms, and dynamic corpora adjustments to reflect evolving regulatory content complexity. All experiments are run on a standard computational server equipped with 32GB RAM and NVIDIA GTX 1080 Ti GPUs. This configuration ensures efficient simulation of complex interactions and supports multiple agent interactions within a single execution cycle.

\subsubsection*{Initial Parameter Estimation using Nonlinear Least Squares}

The optimization process was carried out using \texttt{scipy.optimize.minimize} in Python, employing a nonlinear least squares approach to iteratively refine parameter estimates \cite{ding2023least}. Initial guesses for each parameter were assigned based on preliminary analysis to establish a starting point. The system of differential equations was integrated using \texttt{odeint}, generating model predictions for each variable over the simulation timeline. Parameters were adjusted in each iteration to minimize the objective function, defined as the sum of squared errors between model predictions and observed data for $G_i(t)$, $C_m(t)$, $M_m(t)$, and $F_m^j(t)$ at each time step. The optimization aimed to achieve a minimal objective function value, ensuring a close fit between model outputs and the observed data. Further model validation, including residual analysis and sensitivity testing, was planned to assess the robustness and reliability of the estimated parameters.

To estimate these parameters, we defined an objective function that minimizes the sum of squared errors between model predictions and observed data across all four variables. For each time point $t$, the observed data for $G_i(t)$, $C_m(t)$, $M_m(t)$, and $F_m^j(t)$ was compared to model predictions generated by solving the differential equations with the given parameters.

The objective function was defined as:

\[
\text{Objective} = \sum_t \left( E_G(t) + E_C(t) + E_M(t) + E_F(t) \right)
\]

\[
E_G(t) = \left( G_i(t)_{\text{observed}} - G_i(t)_{\text{predicted}} \right)^2
\]
\[
E_C(t) = \left( C_m(t)_{\text{observed}} - C_m(t)_{\text{predicted}} \right)^2
\]
\[
E_M(t) = \left( M_m(t)_{\text{observed}} - M_m(t)_{\text{predicted}} \right)^2
\]
\[
E_F(t) = \left( F_m^j(t)_{\text{observed}} - F_m^j(t)_{\text{predicted}} \right)^2
\]

\subsection{Evaluation and Validation}
In our study, we employed a comprehensive approach to assess the model's performance and validate its reliability. Evaluation Metrics included metrics such as Guidance Adherence Accuracy and Compliance Stability, chosen for their relevance in measuring how effectively agents align with regulatory requirements and maintain steady compliance efforts. Guidance Adherence Accuracy quantifies the degree to which manufacturer agents’ compliance efforts match regulatory expectations over time, while Compliance Stability evaluates fluctuations in these efforts, reflecting the system’s ability to adapt smoothly to evolving regulations.

To ensure robustness, we applied validation techniques such as sensitivity analysis. By systematically varying model parameters, we examined the impacts on key parameters. This method allowed us to identify critical parameters that significantly influence model dynamics and to confirm that the model’s responses remain consistent and predictable under various scenarios. Additionally, the analysis revealed non-linear and asymmetric effects, demonstrating the model’s capability to capture complex interactions between regulatory guidance and market adaptation.

These measures collectively confirm that our model not only performs well on the test data but also adapts effectively to varying regulatory and market conditions. The combination of precise evaluation and thorough validation ensures that our approach can simulate realistic regulatory environments and agent behaviors with a high degree of reliability. Future work will further enhance this analysis by incorporating real-world data for empirical validation and exploring advanced sensitivity analysis technique.

\textbf{Evaluation Metrics of Guidance Adherence Accuracy}

Guidance adherence measures the degree to which manufacturers align their compliance efforts with the issued guidance. We calculate this as the percentage of time steps in which compliance efforts closely match the recommended values derived from regulatory guidance. For each time step $t$, adherence is determined by comparing the calculated compliance effort $C_m(t)$ with expected effort benchmarks provided by the guidance issuance rate $G_i(t)$:
\[
\text{Adherence Accuracy} = \frac{1}{T} \sum_{t=1}^{T} \mathbb{1}\left( |C_m(t) - G_i(t)| < \epsilon \right)
\]
where $T$ is the total number of time steps, and $epsilon$ is a threshold for acceptable deviation. This metric evaluates how effectively the model influences manufacturers to adopt behaviors in line with regulatory expectations.

 \textbf{Evaluation Metrics of Compliance Stability}

Compliance stability assesses fluctuations in compliance efforts $C_m(t)$ over time, offering insight into the system's resilience to regulatory updates. Lower variance in compliance efforts indicates a smoother adaptation to regulatory changes and reflects an agent's ability to maintain steady compliance:

\[
\text{Compliance Stability} = \frac{1}{T} \sum_{t=1}^{T} \left(C_m(t) - \overline{C_m}\right)^2
\]
where $\overline{C_m}$ is the average compliance effort over $T$ time steps. A lower stability score reflects higher regulatory resilience, suggesting that agents respond consistently even with evolving regulatory requirements.

These metrics collectively allow us to evaluate the regulatory flow model's ability to simulate real-world adaptive behaviors under complex regulatory corpora, providing insight into regulatory effectiveness and agent adaptability in diverse compliance scenarios.

\subsubsection*{Sensitivity Analysis}

We systematically varied each model parameter across a range of values and observed the resulting changes in four key outputs: the Guidance Issuance Rate ($G_i$), Compliance Effort ($C_m$), Market Adaptation ($M_m$), and Manufacturer Feedback ($F_m$). The goal was to assess which parameters exert the most influence on system behavior and to validate the model's structural soundness.

\section{Results} 

\subsection{Regulatory Response and Parameter Evolution of Manufacturer Agents}

\begin{figure}[H]
    \centering
    \includegraphics[width=0.8\textwidth]{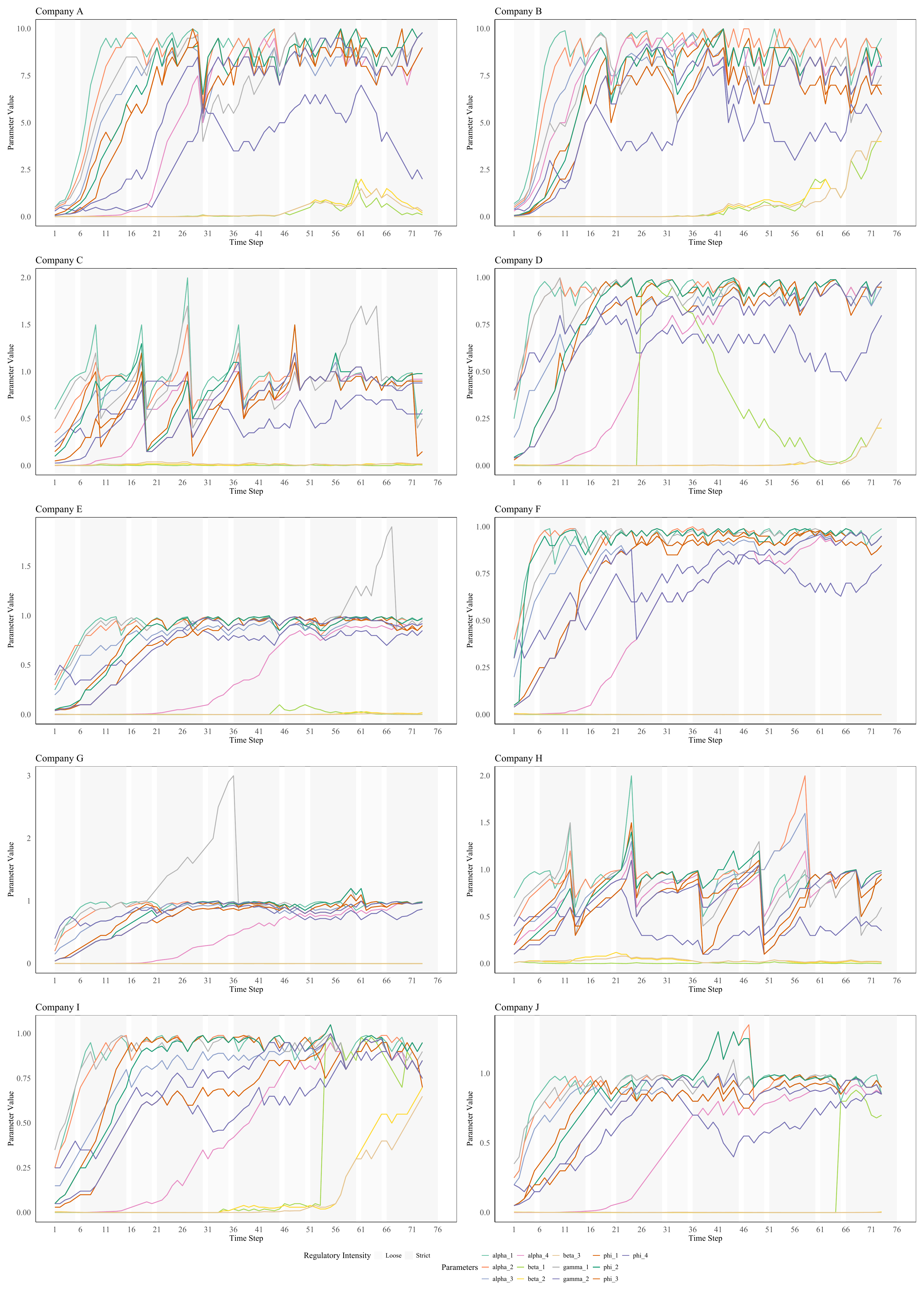}
    \caption{Analysis of Parameter Change and Value Distribution}
    \label{fig:react}
\end{figure}

\begin{figure}[H]
    \centering
    \includegraphics[width=0.8\textwidth]{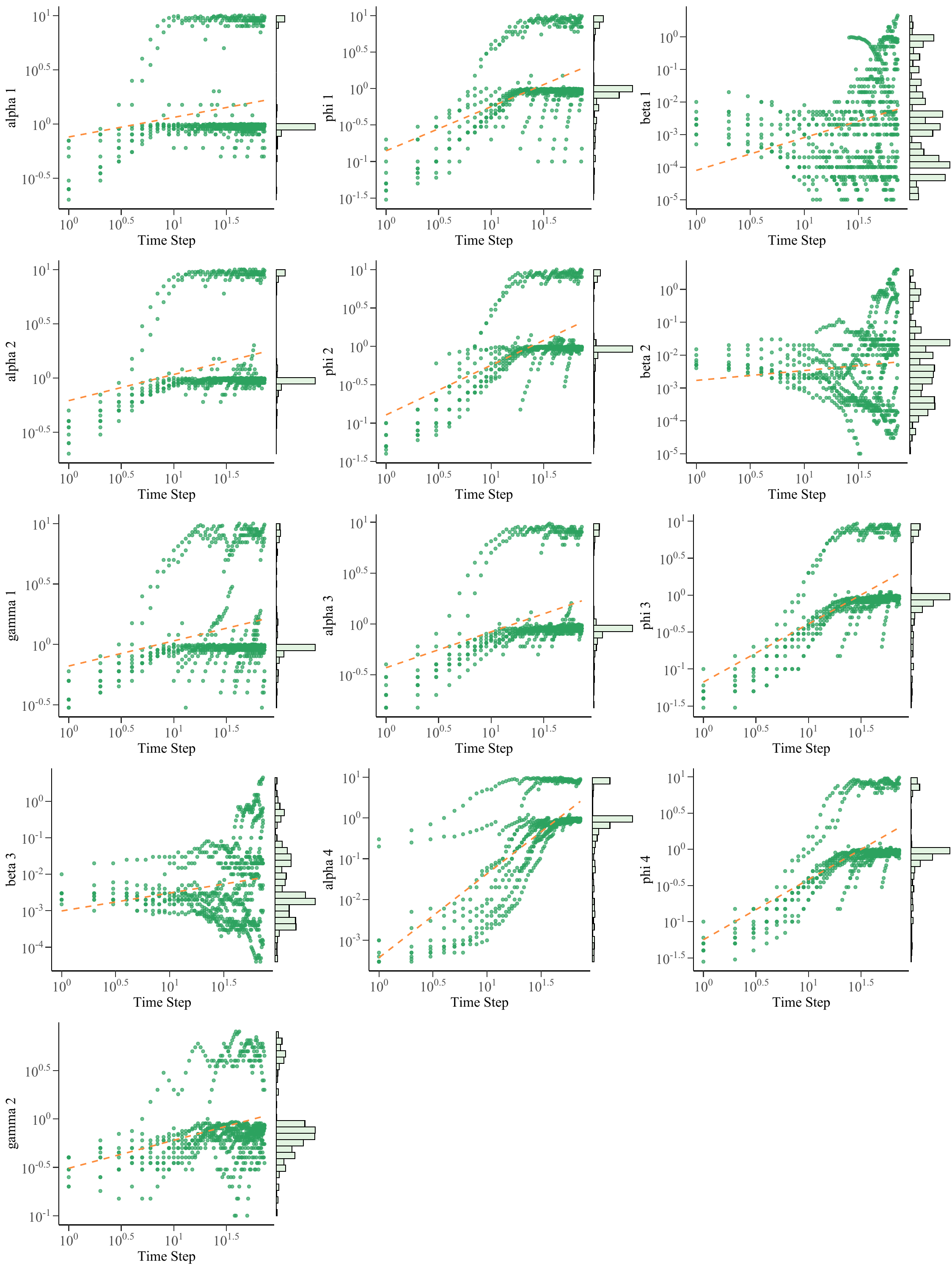}
    \caption{Figure: Analysis of Parameter Sensitivity and Value Distribution}
    \label{fig:pa}
\end{figure}

We conducted a series of experiments in which we subjected different manufacturer agents, to the same regulatory environment. Each agent was provided with identical regulatory guidelines to analyze and respond to. For instance, one of the regulations presented to the agents stated: \textit{"Algorithm Transparency and Traceability: Manufacturers of AI-based medical devices must ensure comprehensive transparency of algorithmic processes. This includes the requirement to document and disclose the decision-making mechanisms at every stage of the model, particularly in complex architectures such as deep neural networks. The traceability of decisions made by the model must be established, providing a clear audit trail of how each layer contributes to the final outcome. This documentation should be structured to enable regulatory bodies to conduct in-depth assessments and identify specific points of failure or risk when necessary."} Upon reading and interpreting the regulations, the agents autonomously decided on strategies to comply, which resulted in adjustments to their operational parameters, such as $\alpha_1, \alpha_2, \beta_1$, and $\gamma_1$, over time as shown in Figure \ref{fig:react} and Figure \ref{fig:pa}. These parameter changes reflect the evolution of their strategies under the imposed regulatory constraints, as shown in the multi-panel visualization of parameter trajectories across companies (A–J). 

\begin{figure}[H]
    \centering
    \includegraphics[width=0.8\textwidth]{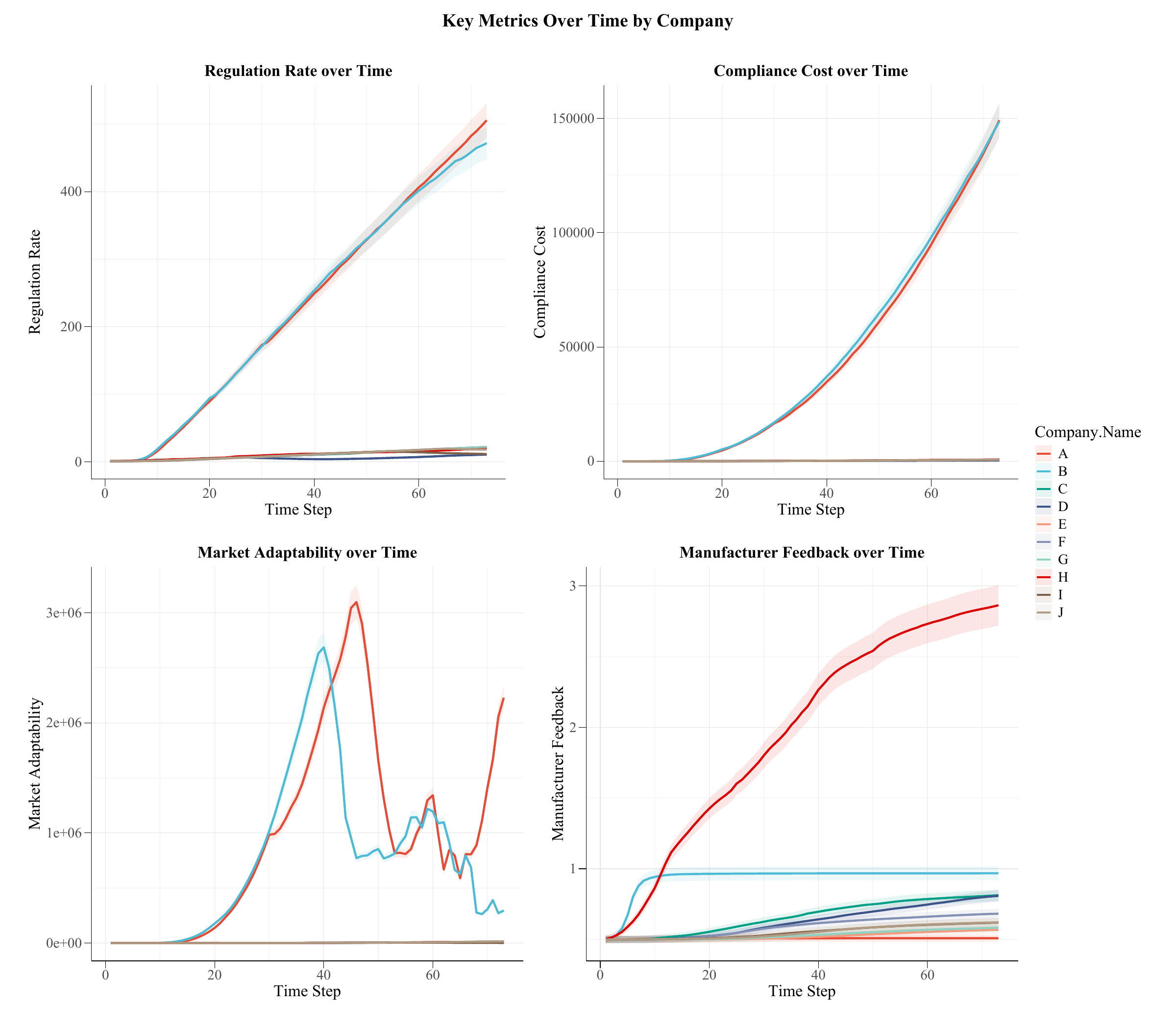}
    \caption{Temporal Dynamics of Key Metrics Under Regulatory Pressure Across Companies}
    \label{fig:value}
\end{figure}

Figure \ref{fig:value} illustrates the temporal dynamics of four key metrics—Regulation Rate, Compliance Cost, Market Adaptability, and Manufacturer Feedback—across multiple companies (A–J) over a series of time steps. The regulation rate (top left) shows a steady and consistent increase across all companies, with some exhibiting faster growth trajectories, indicating variability in their ability to respond to regulatory pressures. Similarly, compliance cost (top right) follows an exponential growth pattern, with companies converging on comparable trajectories over time, reflecting a shared burden of regulatory adherence as the regulation rate increases. In contrast, market adaptability (bottom left) displays significant inter-company variability, with certain companies (e.g., B and D) exhibiting sharp peaks and fluctuations, likely reflecting strategic experimentation or dynamic market responses, while others maintain lower or more stable levels, suggesting resource constraints or differing strategies. Manufacturer feedback (bottom right) grows steadily across companies, with some (e.g., Company J) showing a notably sharper increase, highlighting varying levels of engagement with feedback mechanisms and their potential influence on compliance and adaptability strategies. Together, these metrics provide a comprehensive view of how companies balance regulatory compliance, cost, market adaptability, and external feedback under evolving regulatory environments.

We can see upon exposure to the regulations, each agent processes the requirements and adapts its strategy by modifying internal parameters to achieve compliance while maintaining competitive efficiency.  Initially, during the early time steps, most agents display a rapid upward adjustment in $\alpha_1$ and $\alpha_2$, reflecting a focused effort toward regulatory adherence. This phase is particularly evident in agents representing Companies A, B, and J, where stabilization occurs quickly, suggesting robust initial responses. In the middle time steps, divergence emerges as agents prioritize different strategies: some allocate resources to traceability ($\gamma_1, \gamma_2$), while others focus on market adaptation ($\beta_1, \beta_2$). By the later stages, many agents reach stabilization, balancing regulatory compliance and operational constraints, although some agents, such as those representing Companies C, G, and H, exhibit periodic fluctuations in $\gamma_1$ and $\gamma_2$, indicating ongoing refinements or challenges. Anomalies, such as the sudden drop in $\alpha_2$ for Company G or irregular spikes in $\gamma_1$ for Company D, suggest strategic failures or difficulties in meeting specific requirements, such as establishing audit trails for complex neural networks. These findings highlight the diverse strategies agents adopt, reflecting trade-offs between regulatory compliance and competitiveness.

\subsection{Effect of Resource Constraints on Adaptive Strategies}

\begin{figure}[H]
    \centering
    \includegraphics[width=0.8\textwidth]{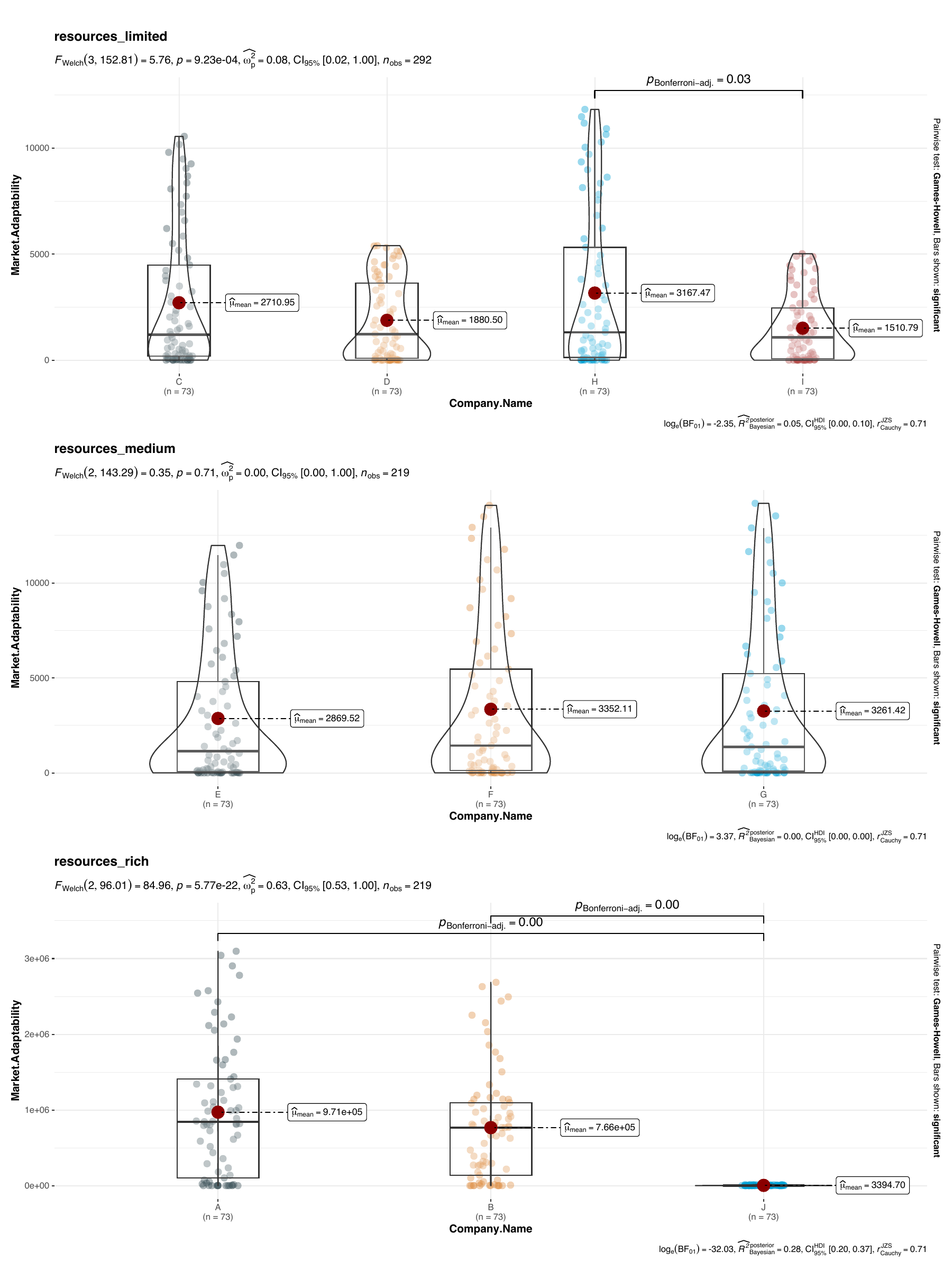}
    \caption{Resource Allocation and Market Adaptability Categorization Across Companies}
    \label{fig:violin}
\end{figure}

This study examines the allocation of compliance resources by each company and its subsequent impact on market adaptability. Based on the results, companies are categorized into three distinct groups reflecting their resource levels: limited, medium, and rich. This categorization allows for a detailed analysis of how varying levels of resource investment influence the ability of companies to adapt to market demands within a regulated environment. The results are visualized using violin plots in Figure \ref{fig:violin}, which are overlaid with summary statistics to display the distribution of market adaptability scores and provide insights into inter-company variability under each condition.

Company C, D, H and I are categorized as \textit{resources\_limited} scenario, significant differences in market adaptability were observed among companies ($F(3, 152.81) = 5.76$, $p = 9.23 \times 10^{-4}$). Company H exhibited the highest mean adaptability ($\bar{x} = 3167.47$), markedly outperforming its counterparts. The Bonferroni-adjusted $p$-value of 0.03 further confirmed the robustness of these findings, while the variance explained by group differences ($\sigma_B^2 = 0.08$) indicates that approximately 8\% of the variability in market adaptability can be attributed to differences among companies. This suggests that under conditions of resource scarcity, certain companies may be better equipped to adapt due to intrinsic factors such as efficiency, innovation, or strategic management.

The \textit{resources\_medium} scenario, Agent E, F, G, revealed no statistically significant differences in market adaptability among companies ($F(2, 143.29) = 0.35$, $p = 0.71$). The mean adaptability scores across all companies were closely aligned, indicating that moderate levels of resources may result in homogenized performance. The explained variance was effectively negligible ($\sigma_B^2 = 0.00$), suggesting that adaptability under these conditions is largely independent of company-level factors. This observation highlights the stabilizing effects of balanced resource availability, which may reduce inter-company disparities.

The \textit{resources\_rich} scenario, Agent A, B and J, on the other hand, displayed pronounced differences in market adaptability across companies ($F(2, 96.01) = 84.96$, $p = 5.77 \times 10^{-22}$). Company A demonstrated an extraordinary mean adaptability score ($\bar{x} = 9.71 \times 10^5$), significantly exceeding the performance of other companies. The variance explained by group differences was substantial ($\sigma_B^2 = 0.63$), indicating that approximately 63\% of the variability in market adaptability can be attributed to company identity under resource-rich conditions. The Bonferroni-adjusted $p$-values ($p < 0.001$) confirmed the significance of these differences, reinforcing the notion that abundant resources magnify inter-company disparities, possibly by amplifying existing advantages in technology, infrastructure, or market reach.

The findings illustrate a complex relationship between resource availability and market adaptability. Under limited resources, adaptability disparities are moderate but significant, likely driven by intrinsic company characteristics. Medium resource conditions neutralize these differences, suggesting a stabilizing effect of balanced resource allocation. However, resource abundance appears to exacerbate adaptability disparities, with certain companies achieving extraordinary levels of performance. These results underscore the importance of understanding how resource availability interacts with company-level factors to influence competitive dynamics and market success.

Figure \ref{fig:sen} illustrates the market adaptability achieved by each company after 73 time steps, highlighting the outcomes of their strategic decisions and compliance resource allocations. The vertical axis represents market adaptability, while the horizontal axis categorizes companies (A–J). Companies within the resources\_rich category (e.g., A, B, and J) consistently achieve the highest adaptability scores, underscoring the advantages of abundant resources in fostering strategic flexibility and responsiveness to market dynamics. In contrast, companies categorized as resources\_limited (e.g., C, D, H, and I) exhibit significantly lower adaptability, with notable variability among them. Notably, Company H outperforms its peers within this category, suggesting superior efficiency or innovative strategies despite resource constraints. Further investigation into Company H reveals that its higher market adaptability is driven by its strategic focus on AI integration in preventive care devices, which aligns with its innovative product portfolio and the increasing demand for advanced healthcare solutions. Additionally, its low-to-medium risk preference facilitates targeted AI investments (6\% of revenue) while maintaining operational stability, enabling it to achieve greater adaptability even under resource-limited conditions.

  \begin{figure}[H]
    \centering
    \includegraphics[width=1\textwidth]{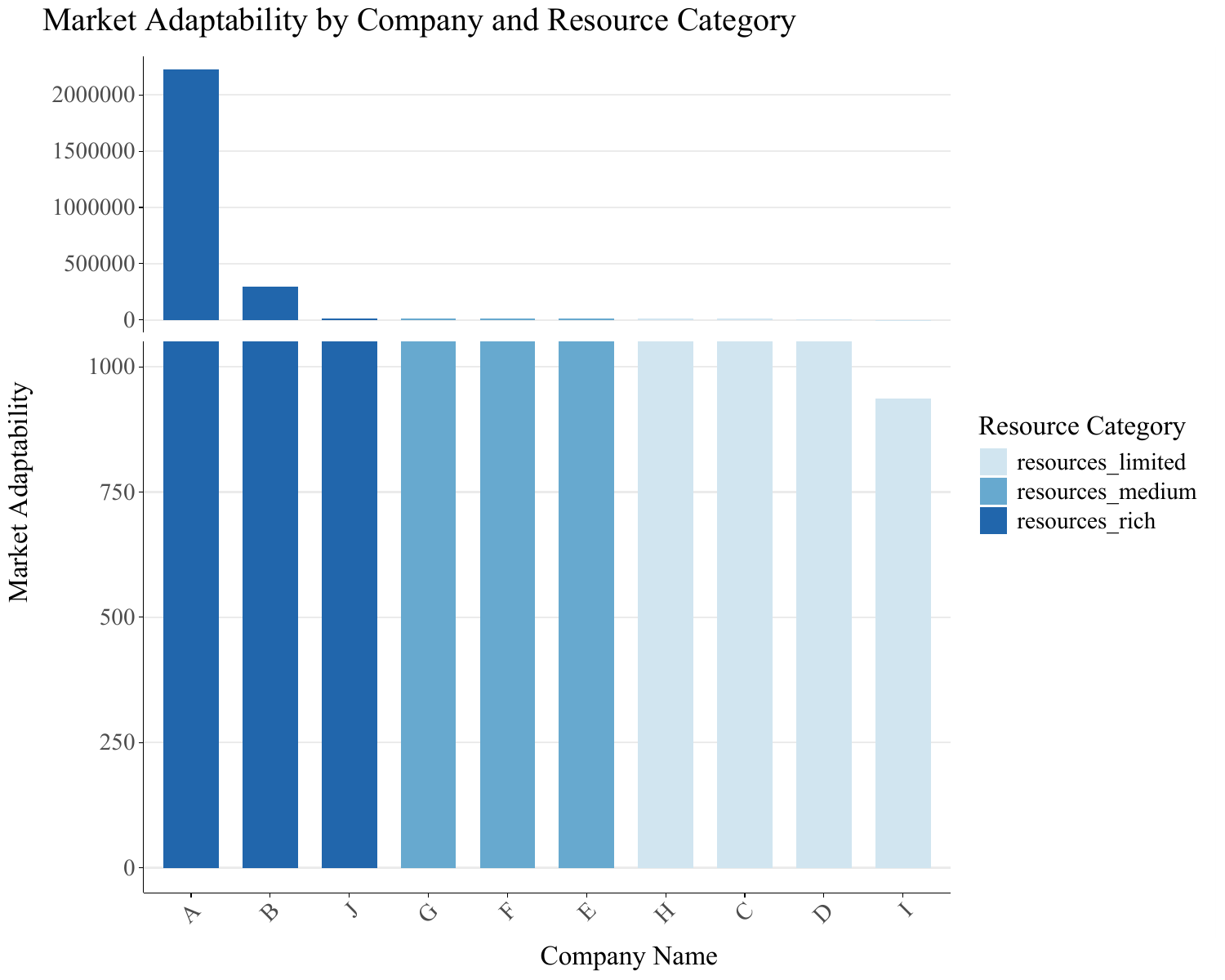}
    \caption{Analysis of Market Adaptability Value Distribution}
    \label{fig:sen}
\end{figure}

\subsection{Decision made by Authority Agent}

The "Benefit-Risk Ratio" (BRR) serves as a foundational metric in regulatory decision-making, enabling authorities to assess whether a product's potential benefits outweigh its associated risks sufficiently to justify market entry or continued presence. This ratio acts as a quantitative and qualitative guide for balancing innovation and safety, especially in highly regulated industries such as healthcare, where public safety and efficacy are paramount. By analyzing BRR, regulatory bodies can evaluate complex scenarios involving diverse manufacturers, varying product types, and evolving market dynamics.

The authority agent take the BRR tool allows the authority to assess the overall market landscape, including the number of manufacturers present and the diversity of their products. Based on this comprehensive review, the authority agent sets a decision threshold to determine the acceptable level of benefit relative to risk for products entering or remaining in the market. Unlike real-world regulatory practices, where decisions are made on a case-by-case basis with significant flexibility for individual product characteristics, the authority agent in this model introduces a dynamic thresholding mechanism.

\begin{figure}[H]
    \centering
    \includegraphics[width=0.8\textwidth]{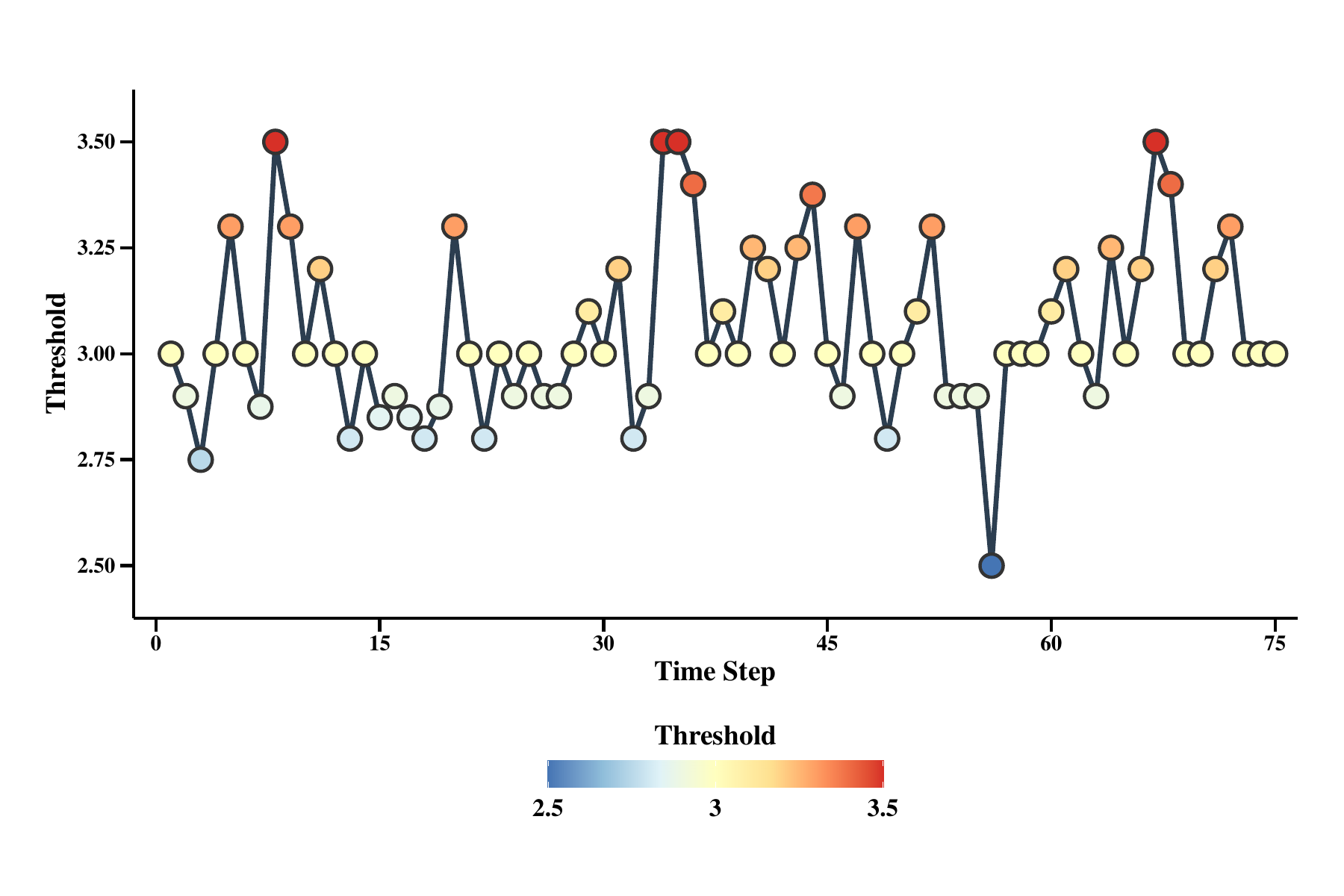}
    \caption{BRR value change set by multi agent framework}
    \label{fig:calculator}
\end{figure}

As shown in Figure \ref{fig:calculator}, the threshold set by the authority agent is not a fixed value but fluctuates over time, reflecting adaptive decision-making based on changing market conditions. The fluctuation in the threshold ensures responsiveness to variations in market dynamics, such as the entry of new manufacturers or changes in product benefit-risk profiles. This dynamic approach mirrors real-life regulatory flexibility, where decisions often account for context-specific factors and evolving market realities. By analyzing the threshold trajectory, we observe a balance between maintaining rigorous standards and allowing flexibility to accommodate innovation and market diversity. The adaptive nature of the authority’s decision-making underscores the importance of balancing regulatory consistency with the need for situational responsiveness in complex and competitive markets.

\subsection{Module evaluation and sensitivity result}
To validate the robustness and relevance of the framework, we conducted a comprehensive sensitivity analysis. The results are summarized and depicted in Figure \ref{fig:sen}, showcasing the change rates for each key parameter across critical outputs: Guidance Issuance Rate ($G_i$), Compliance Effort ($C_m$), Market Adaptation ($M_m$), and Manufacturer Feedback ($F_m$).

\begin{figure}[H]
    \centering
    \includegraphics[width=0.9\textwidth]{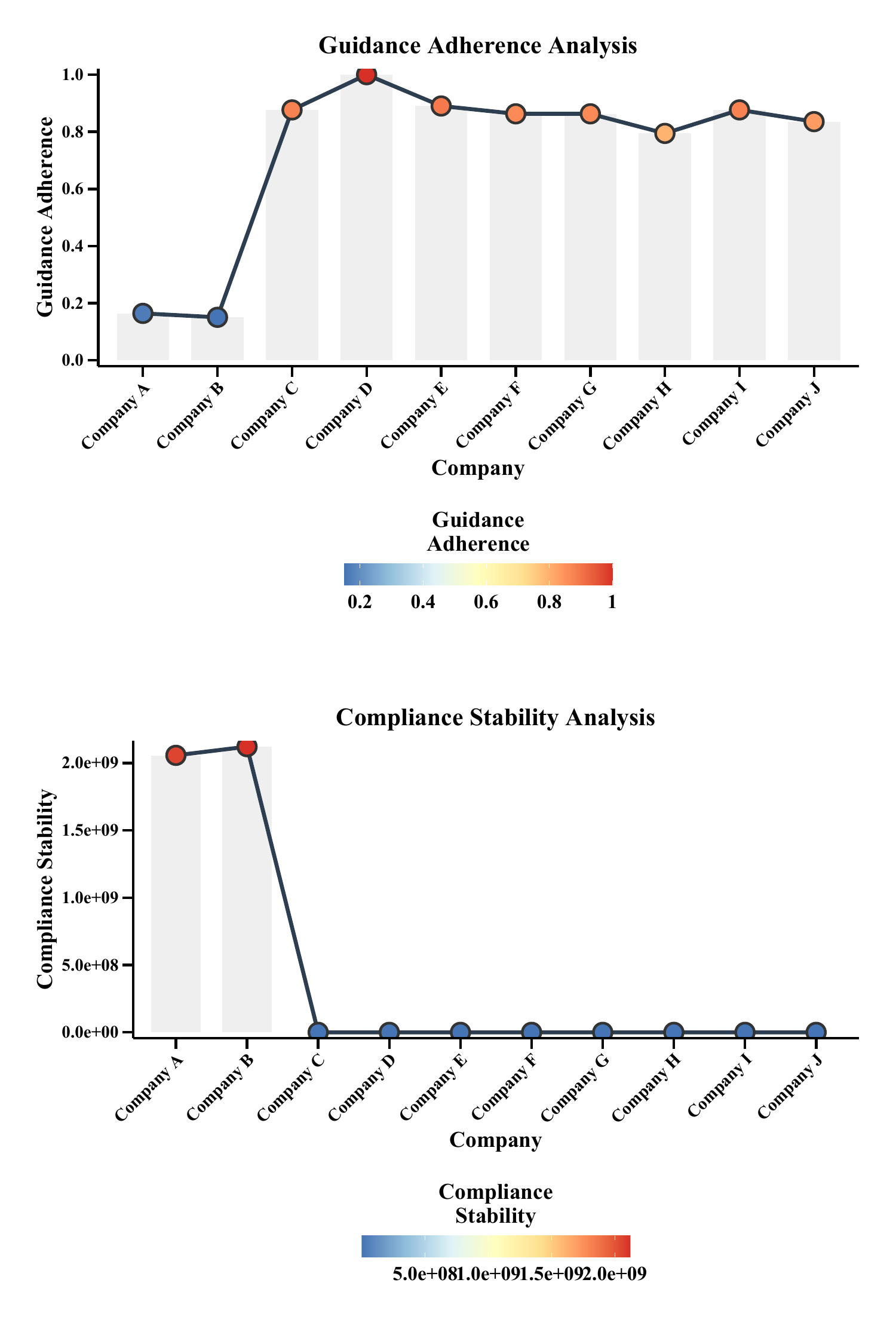}
    \caption{Evaluation result of Guidance adherence analysis and compliance stability analysis}
    \label{fig:sen}
\end{figure}

  \begin{figure}[H]
    \centering
    \includegraphics[width=1\textwidth]{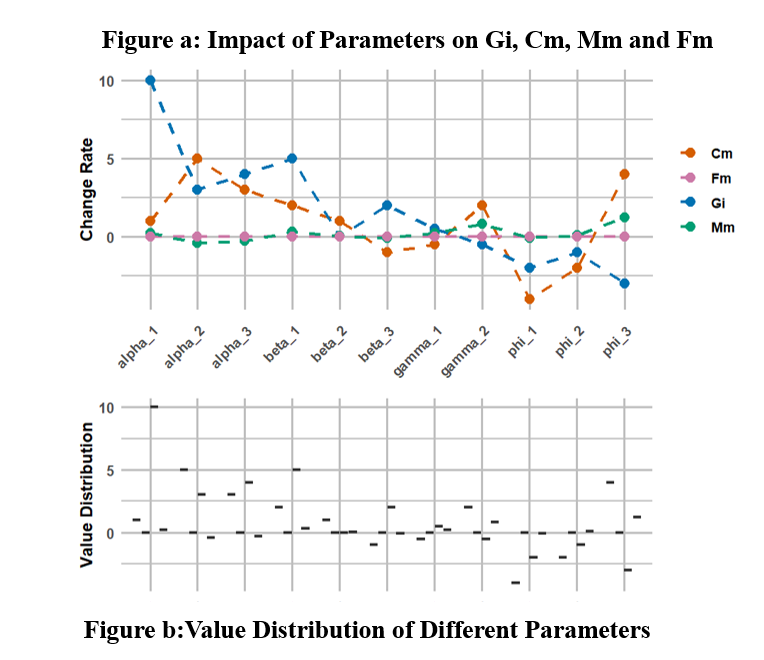}
    \caption{Figure: Analysis of Parameter Sensitivity and Value Distribution}
    \label{fig:sen}
\end{figure}

\subsection{Sensitive analysis}

\paragraph{Impact of Parameters on Model Outputs}

The parameters $\alpha_1$, $\phi_1$, and $\beta_1$, which govern the Guidance Issuance Rate, exhibited a moderate to strong influence on $G_i$, with changes leading to both positive and negative shifts in output. Notably, increasing $\alpha_1$ from 0.1 to 0.2 resulted in significant increases in $G_i$, suggesting that regulatory guidance is sensitive to changes in the maximum issuance rate. Parameters such as $\phi_2$ and $\beta_2$, which influence Compliance Effort ($C_m$), demonstrated non-linear impacts. For instance, as $\phi_2$ increased, there was a notable positive effect on $C_m$, indicating that research and development investments have a pronounced impact on compliance behavior.

The sensitivity of Market Adaptation ($M_m$) was most pronounced with variations in $\alpha_3$ and $\beta_3$. Specifically, higher values of $\alpha_3$ accelerated the rate at which compliance efforts translated into market adaptation, confirming the importance of resource allocation in this aspect. Feedback from manufacturers ($F_m$) was shown to be moderately influenced by adjustments in $\alpha_4$ and $\phi_4$, which indicates the role of market adaptation and compliance capability in shaping feedback dynamics.

\paragraph{Critical Parameters and Their Influence}
Parameters like $\alpha_1$, $\alpha_2$, and $\alpha_3$ were identified as critical, with relatively large change rates in key outputs when adjusted. This finding emphasizes the importance of accurately estimating these parameters, as they significantly drive the regulatory flow and market response dynamics. Conversely, some parameters, such as $\gamma_1$ and $\gamma_2$, showed more stable behavior, suggesting lower sensitivity. This implies that while these parameters are still relevant, their precise values may be less crucial for overall model performance, which could simplify parameter estimation.

\paragraph{Non-linear and Asymmetric Effects}
The analysis revealed that the effects of parameter changes are often non-linear and asymmetric. For example, increasing $\beta_1$ resulted in diminishing marginal impacts on $G_i$, demonstrating a saturation effect. Similarly, variations in $\phi_3$ had asymmetric impacts on $M_m$, which highlights the need for careful tuning to avoid disproportionate responses.

The results of our sensitivity analysis validate the robustness of our model, as it consistently exhibited expected and explainable behaviors across a wide range of parameter variations. The identification of critical parameters underscores the importance of focusing on these variables during calibration and further analysis. Additionally, the non-linear responses observed suggest that our model captures the complex dynamics of regulatory and market interactions effectively.

While the sensitivity analysis provides valuable insights, it does not replace empirical validation against real-world data. Future work should include extensive data collection to refine parameter estimates and validate model predictions. Additionally, exploring more advanced sensitivity analysis techniques, such as global sensitivity analysis, could provide a more comprehensive understanding of model dynamics.


\section{Discussion}

This study introduces a multi-agent framework designed to model regulatory compliance and market adaptation dynamics in the medical device industry. By simulating the interactions between regulatory bodies and manufacturers, the framework provides a novel lens to examine policy impacts, strategic decision-making, and adaptive behaviors. Despite its utility, the model entails certain limitations that warrant further investigation and refinement to enhance its robustness and applicability in real-world settings.

Regulatory policy analysis could benefit significantly from this modeling approach. Regulatory bodies might employ the model to simulate the potential impact of new policies on the medical device industry. By adjusting parameters such as $\alpha_1$, $\beta_1$, and $\phi_1$ in Equation (1), regulators could model scenarios involving varying guidance rates and their effects on compliance efforts and market adaptation. While this predictive capability could enable more informed policy-making, ensuring that regulations are both effective and balanced, the accuracy of these predictions depends heavily on the quality and completeness of input data, which is a limitation in real-world applications.

The model holds potential for strategic planning for manufacturers. Medical device manufacturers could leverage the framework to anticipate regulatory changes and adjust compliance strategies proactively. By inputting specific data—such as compliance scores, R\&D investments, and market share—the model might enable companies to forecast future compliance costs and evaluate market adaptation strategies. However, the variability of real-world conditions, such as sudden regulatory changes or market disruptions, limits the precision of these forecasts, making the model better suited for strategic guidance rather than exact predictions. It also serves as a basis for academic Research and Validation. Researchers could explore the broader implications of regulatory impacts within the medtech industry by using this framework. Sensitivity analyses and validations using real-world data could provide valuable insights into its robustness and applicability. However, the model’s current limitations, such as assumptions of uniform compliance behavior or simplified market dynamics, must be addressed to improve its credibility and generalizability. The model aspires to support industry benchmarking. By comparing predicted compliance efforts and market adaptation outcomes with actual industry performance, companies could identify performance gaps and opportunities for improvement. This process might facilitate the establishment of industry standards and promote continuous improvement. However, the model’s inherent simplifications might limit its ability to capture the nuanced, firm-specific factors that drive real-world performance, reducing its effectiveness as a benchmarking tool in its current state.

\section{Limitation}
The primary limitation of this study stems from the simplifications inherent to the BRR (Benefit Risk Ratio) scoring system, which leverages large language model to assign scores based on manufacturers' profiles, device descriptions, and regulatory environments. While the BRR system employs logical principles grounded in regulatory decision-making processes, it remains a proxy for real-world evaluations, which are inherently more complex and multifaceted.

In practice, the BRR consideration and final approval of medical devices are influenced by a wide range of additional factors, including clinical trial results, post-market surveillance data, performance in animal studies, laboratory testing outcomes, and urgent medical needs. For example, evaluating a cardiac implantable device in real-world conditions requires assessing its long-term clinical effectiveness, biocompatibility, and cost-efficiency, as well as its alignment with reimbursement policies and patient-centered outcomes. These complexities go beyond the scope of the current model, which relies mainly on structured input features and generalized decision-making rules.


This study represents a foundational step in developing a quantitative framework for modeling regulatory compliance and market adaptation. As a first attempt, it necessarily abstracts away certain product-specific and firm-specific details to maintain computational tractability and generalizability. Future work can build on this foundation by integrating individual attributes of the granular product and real-world data sources to improve the accuracy and scope of the model.


\section{Conclusion and Future Work} \label{Section: conclusion}

In this paper, we introduced a novel simulation multi agent LLM framework designed to evaluate and enhance regulatory compliance strategies for AI-based medical device manufacturers. By representing both manufacturers and regulatory bodies, such as the FDA, as autonomous agents powered by large language models, this framework enables iterative, interactive compliance assessments that mimic real-world regulatory processes. Our approach provides a unique solution to address the regulatory complexities associated with the adaptive nature of AI in medical devices, where compliance is a dynamic and ongoing challenge. The metrics of this system, including compliance accuracy, stability, and compliance with regulatory guidance, allow a multidimensional view of compliance performance. Experimental results demonstrated that this iterative learning environment enables manufacturers to continuously refine compliance strategies, while regulatory agents can provide tailored feedback to guide manufacturers toward optimal compliance. Such a setup not only reduces potential compliance risks but also promotes faster and more efficient market readiness for innovative AI-based medical devices. Beyond its immediate application in medical device compliance, this framework has broader potential in various high-regulation industries, including financial technology, biopharmaceuticals, and autonomous vehicles. It represents a significant step forward in simulating and analyzing regulatory compliance for AI-driven technologies. It lays the groundwork for proactive compliance assessment, fosters better alignment with regulatory standards, and enhances safety and efficacy in AI applications. As such, it offers a promising pathway for advancing the role of AI-powered agent systems in regulatory science, while opening avenues for cross-disciplinary research and collaboration across sectors that prioritize safety, compliance, and innovation.

\bibliographystyle{IEEEtran}
\bibliography{od_prediction}

\begin{thebibliography}{10}
\providecommand{\url}[1]{#1}
\csname url@samestyle\endcsname
\providecommand{\newblock}{\relax}
\providecommand{\bibinfo}[2]{#2}
\providecommand{\BIBentrySTDinterwordspacing}{\spaceskip=0pt\relax}
\providecommand{\BIBentryALTinterwordstretchfactor}{4}
\providecommand{\BIBentryALTinterwordspacing}{\spaceskip=\fontdimen2\font plus
\BIBentryALTinterwordstretchfactor\fontdimen3\font minus
  \fontdimen4\font\relax}
\providecommand{\BIBforeignlanguage}[2]{{%
\expandafter\ifx\csname l@#1\endcsname\relax
\typeout{** WARNING: IEEEtran.bst: No hyphenation pattern has been}%
\typeout{** loaded for the language `#1'. Using the pattern for}%
\typeout{** the default language instead.}%
\else
\language=\csname l@#1\endcsname
\fi
#2}}
\providecommand{\BIBdecl}{\relax}
\BIBdecl

\bibitem{han2024transforming}
Y.~Han and J.~Bergmann, ``Transforming medical regulations into numbers:
  Vectorizing a decade of medical device regulatory shifts in the usa, eu, and
  china,'' \emph{arXiv preprint arXiv:2411.00567}, 2024.

\bibitem{kramer2012does}
D.~B. Kramer, S.~Xu, and A.~S. Kesselheim, ``How does medical device regulation
  perform in the united states and the european union? a systematic review,''
  2012.

\bibitem{han2024evaluation}
Y.~Han, A.~Ceross, F.~Bourgeois, P.~Savaget, and J.~H. Bergmann, ``Evaluation
  of large language models for the classification of medical device software,''
  2024.

\bibitem{anderson1999perspective}
P.~Anderson, ``Perspective: Complexity theory and organization science,''
  \emph{Organization science}, vol.~10, no.~3, pp. 216--232, 1999.

\bibitem{sterman2000business}
J.~D. Sterman, ``Business dynamics: Systems thinking and modeling for a complex
  world,'' \emph{MacGraw-Hill Company}, 2000.

\bibitem{kauffman1993origins}
S.~A. Kauffman, \emph{The origins of order: Self-organization and selection in
  evolution}.\hskip 1em plus 0.5em minus 0.4em\relax Oxford University Press,
  1993.

\bibitem{Helbing2013HowTD}
\BIBentryALTinterwordspacing
D.~Helbing and S.~Balietti, ``How to do agent-based simulations in the future:
  From modeling social mechanisms to emergent phenomena and interactive systems
  design,'' \emph{Urban Economics \& Regional Studies eJournal}, 2013.
  [Online]. Available: \url{https://api.semanticscholar.org/CorpusID:5752779}
\BIBentrySTDinterwordspacing

\bibitem{lottes2022navigating}
A.~E. Lottes, K.~J. Cavanaugh, Y.~Y.-F. Chan, V.~J. Devlin, C.~J. Goergen,
  R.~Jean, J.~C. Linnes, M.~Malone, R.~Peat, D.~G. Reuter \emph{et~al.},
  ``Navigating the regulatory pathway for medical devices—a conversation with
  the fda, clinicians, researchers, and industry experts,'' \emph{Journal of
  Cardiovascular Translational Research}, vol.~15, no.~5, pp. 927--943, 2022.

\bibitem{wooldridge2009introduction}
M.~Wooldridge, \emph{An introduction to multiagent systems}.\hskip 1em plus
  0.5em minus 0.4em\relax John wiley \& sons, 2009.

\bibitem{parker2003multi}
D.~C. Parker, S.~M. Manson, M.~A. Janssen, M.~J. Hoffmann, and P.~Deadman,
  ``Multi-agent systems for the simulation of land-use and land-cover change: a
  review,'' \emph{Annals of the association of American Geographers}, vol.~93,
  no.~2, pp. 314--337, 2003.

\bibitem{liu2024multi}
L.~Liu, M.~Wang, M.-O. Pun, and X.~Xiong, ``A multi-agent rollout approach for
  highway bottleneck decongenston in mixed autonomy,'' \emph{arXiv preprint
  arXiv:2405.03132}, 2024.

\bibitem{xie2024large}
J.~Xie, Z.~Chen, R.~Zhang, X.~Wan, and G.~Li, ``Large multimodal agents: A
  survey,'' \emph{arXiv preprint arXiv:2402.15116}, 2024.

\bibitem{guo2024large}
T.~Guo, X.~Chen, Y.~Wang, R.~Chang, S.~Pei, N.~V. Chawla, O.~Wiest, and
  X.~Zhang, ``Large language model based multi-agents: A survey of progress and
  challenges,'' \emph{arXiv preprint arXiv:2402.01680}, 2024.

\bibitem{turner2019complexity}
J.~R. Turner and R.~M. Baker, ``Complexity theory: An overview with potential
  applications for the social sciences,'' \emph{Systems}, vol.~7, no.~1, p.~4,
  2019.

\bibitem{macal2016everything}
C.~M. Macal, ``Everything you need to know about agent-based modelling and
  simulation,'' \emph{Journal of Simulation}, vol.~10, pp. 144--156, 2016.

\bibitem{axelrod2008harnessing}
R.~Axelrod and M.~D. Cohen, \emph{Harnessing complexity}.\hskip 1em plus 0.5em
  minus 0.4em\relax Basic books, 2008.

\bibitem{hedstrom2010causal}
P.~Hedstr{\"o}m and P.~Ylikoski, ``Causal mechanisms in the social sciences,''
  \emph{Annual review of sociology}, vol.~36, pp. 49--67, 2010.

\bibitem{wang2024reinforcement}
Y.~Wang, L.~Liu, M.~Wang, and X.~Xiong, ``Reinforcement learning from human
  feedback for lane changing of autonomous vehicles in mixed traffic,''
  \emph{arXiv preprint arXiv:2408.04447}, 2024.

\bibitem{turner2011editor}
J.~R. Turner, ``Editor's commentary: regulatory science and the science of
  safety,'' \emph{Drug Information Journal}, vol.~45, no.~3, pp. 221--227,
  2011.

\bibitem{han2024more}
Y.~Han, A.~Ceross, and J.~Bergmann, ``More than red tape: exploring complexity
  in medical device regulatory affairs,'' \emph{Frontiers in Medicine},
  vol.~11, p. 1415319, 2024.

\bibitem{han2024revolutionizing}
Y.~Han and J.~Tao, ``Revolutionizing pharma: Unveiling the ai and llm trends in
  the pharmaceutical industry,'' \emph{arXiv preprint arXiv:2401.10273}, 2024.

\bibitem{han2024use}
Y.~Han, A.~Ceross, and J.~H.~M. Bergmann, ``The use of readability metrics in
  legal text: A systematic literature review,'' 2024.

\bibitem{stillhart2019pbpk}
C.~Stillhart, X.~Pepin, C.~Tistaert, D.~Good, A.~Van Den~Bergh, N.~Parrott, and
  F.~Kesisoglou, ``Pbpk absorption modeling: establishing the in vitro--in vivo
  link—industry perspective,'' \emph{The AAPS Journal}, vol.~21, pp. 1--13,
  2019.

\bibitem{morrison2018advancing}
T.~M. Morrison, P.~Pathmanathan, M.~Adwan, and E.~Margerrison, ``Advancing
  regulatory science with computational modeling for medical devices at the
  fda's office of science and engineering laboratories,'' \emph{Frontiers in
  medicine}, vol.~5, p. 241, 2018.

\bibitem{o2019ispor}
T.~O'Neill, R.~Miksad, D.~Miller, L.~Maloney, A.~John, C.~Hiller, and
  J.~Hornberger, ``Ispor, the fda, and the evolving regulatory science of
  medical device products,'' \emph{Value in Health}, vol.~22, no.~7, pp.
  754--761, 2019.

\bibitem{wang2018analytical}
Y.~Wang, A.~Guan, S.~Wickramasekara, and K.~S. Phillips, ``Analytical chemistry
  in the regulatory science of medical devices,'' \emph{Annual Review of
  Analytical Chemistry}, vol.~11, no.~1, pp. 307--327, 2018.

\bibitem{ICH2005}
{International Conference on Harmonisation of Technical Requirements for
  Registration of Pharmaceuticals for Human Use}, ``{ICH Harmonised Tripartite
  Guideline: Quality Risk Management Q9},'' {International Conference on
  Harmonisation of Technical Requirements for Registration of Pharmaceuticals
  for Human Use}, {Geneva, Switzerland}, Tech. Rep., 2005.

\bibitem{fda2006guidance}
U.~FDA, ``Guidance for industry: Q10 quality systems approach to pharmaceutical
  cgmp regulations,'' \emph{Rockville, MD}, 2006.

\bibitem{gonccalves2020risk}
M.~E. Gon{\c{c}}alves, ``The risk-based approach under the new eu data
  protection regulation: a critical perspective,'' \emph{Journal of Risk
  Research}, vol.~23, no.~2, pp. 139--152, 2020.

\bibitem{smith2017structured}
M.~Y. Smith, I.~Benattia, C.~Strauss, L.~Bloss, and Q.~Jiang, ``Structured
  benefit-risk assessment across the product lifecycle: practical
  considerations,'' \emph{Therapeutic Innovation \& Regulatory Science},
  vol.~51, no.~4, pp. 501--508, 2017.

\bibitem{weinstein1977foundations}
M.~C. Weinstein and W.~B. Stason, ``Foundations of cost-effectiveness analysis
  for health and medical practices,'' \emph{New England journal of medicine},
  vol. 296, no.~13, pp. 716--721, 1977.

\bibitem{drummond2015methods}
M.~F. Drummond, M.~J. Sculpher, K.~Claxton, G.~L. Stoddart, and G.~W. Torrance,
  \emph{Methods for the economic evaluation of health care programmes}.\hskip
  1em plus 0.5em minus 0.4em\relax Oxford university press, 2015.

\bibitem{xiong2019analysis}
X.~Xiong, E.~Xiao, and L.~Jin, ``Analysis of a stochastic model for coordinated
  platooning of heavy-duty vehicles,'' in \emph{2019 IEEE 58th conference on
  decision and control (CDC)}.\hskip 1em plus 0.5em minus 0.4em\relax IEEE,
  2019, pp. 3170--3175.

\bibitem{bao2024dynamic}
Y.~Bao, N.~Buhay, and Q.~Zheng, ``A dynamic model for gmp compliance and
  regulatory science,'' \emph{Journal of Pharmaceutical Innovation}, vol.~19,
  no.~3, pp. 1--16, 2024.

\bibitem{gao2024empowering}
S.~Gao, A.~Fang, Y.~Huang, V.~Giunchiglia, A.~Noori, J.~R. Schwarz,
  Y.~Ektefaie, J.~Kondic, and M.~Zitnik, ``Empowering biomedical discovery with
  ai agents,'' \emph{Cell}, vol. 187, no.~22, pp. 6125--6151, 2024.

\bibitem{kurzinger2020structured}
M.-L. K{\"u}rzinger, L.~Douarin, I.~Uzun, C.~El-Haddad, W.~Hurst, J.~Juhaeri,
  and S.~Tcherny-Lessenot, ``Structured benefit--risk evaluation for medicinal
  products: review of quantitative benefit--risk assessment findings in the
  literature,'' \emph{Therapeutic Advances in Drug Safety}, vol.~11, p.
  2042098620976951, 2020.

\bibitem{li2023camel}
G.~Li, H.~Hammoud, H.~Itani, D.~Khizbullin, and B.~Ghanem, ``Camel:
  Communicative agents for" mind" exploration of large language model
  society,'' \emph{Advances in Neural Information Processing Systems}, vol.~36,
  pp. 51\,991--52\,008, 2023.

\bibitem{verhulst1838law}
P.~F. Verhulst, \emph{Notice sur la loi que la population suit dans son
  accroissement}.\hskip 1em plus 0.5em minus 0.4em\relax Correspondance
  Math{\'e}matique et Physique, 1838.

\bibitem{strogatz2018nonlinear}
S.~H. Strogatz, \emph{Nonlinear Dynamics and Chaos: With Applications to
  Physics, Biology, Chemistry, and Engineering}.\hskip 1em plus 0.5em minus
  0.4em\relax CRC Press, 2018.

\bibitem{rogers2003diffusion}
E.~M. Rogers, \emph{Diffusion of Innovations}.\hskip 1em plus 0.5em minus
  0.4em\relax Free Press, 2003.

\bibitem{forrester1968principles}
J.~W. Forrester, \emph{Principles of Systems}.\hskip 1em plus 0.5em minus
  0.4em\relax MIT Press, 1968.

\bibitem{murray2002mathematical}
J.~D. Murray, \emph{Mathematical Biology I: An Introduction}.\hskip 1em plus
  0.5em minus 0.4em\relax Springer, 2002.

\bibitem{ding2023least}
F.~Ding, ``Least squares parameter estimation and multi-innovation least
  squares methods for linear fitting problems from noisy data,'' \emph{Journal
  of Computational and Applied Mathematics}, vol. 426, p. 115107, 2023.

\end{thebibliography}
\end{document}